\documentclass{article}

\usepackage{PRIMEarxiv}

\usepackage[utf8]{inputenc} 
\usepackage[T1]{fontenc}    
\usepackage{hyperref}       
\usepackage{url}            
\usepackage{booktabs}       
\usepackage{amsfonts}       
\usepackage{nicefrac}       
\usepackage{microtype}      
\usepackage{lipsum}
\usepackage{fancyhdr}       
\usepackage{graphicx}       
\graphicspath{{figures/}{media/}}
\usepackage{xcolor}

\usepackage[labelfont=bf]{caption}

\usepackage{enumitem}      

\setlist{nosep,leftmargin=*}

\usepackage{algorithm}
\usepackage{algpseudocode} 
\usepackage{amsmath}
\usepackage{xspace}
\usepackage{float}

\newcommand{\sclaw}{\textsc{ScienceClaw}\xspace}
\newcommand{\ifinf}{\textsc{Infinite}\xspace}

\title{
Autonomous Agents Coordinating Distributed Discovery Through Emergent Artifact Exchange
\thanks{\textit{\underline{Citation}}: 
\textbf{Authors. Title. Pages.... DOI:000000/11111.}} 
}

\author{
  Fiona Y. Wang\textsuperscript{1, 2} \quad
  Lee Marom\textsuperscript{1, 3} \quad
  Subhadeep Pal\textsuperscript{1, 4} \quad
  Rachel K. Luu\textsuperscript{1, 5} \quad
  Wei Lu\textsuperscript{1, 4} \quad \\
  \textbf{Jaime A. Berkovich\textsuperscript{1, 5}} \quad
  \textbf{Markus J. Buehler\textsuperscript{1, 3, 4, 6, \#}} \\ \\
  \textsuperscript{1}Laboratory for Atomistic and Molecular Mechanics (LAMM) \\
  \textsuperscript{2}Department of Biological Engineering \\
  \textsuperscript{3}Department of Mechanical Engineering \\
  \textsuperscript{4}Department of Civil and Environmental Engineering \\
  \textsuperscript{5}Department of Materials Science and Engineering \\
  \textsuperscript{6}Center for Computational Science and Engineering, Schwarzman College of Computing \\
  Massachusetts Institute of Technology, Cambridge, MA 02139, USA \\ \\
  \textsuperscript{\#} Corresponding author: 
  \texttt{mbuehler@MIT.EDU}
}

\begin{document}
\maketitle

\begin{abstract}
We present \sclaw + \ifinf, a framework for autonomous scientific investigation in which independent agents conduct research without central coordination, and any contributor can deploy new agents into a shared ecosystem. The system is built around three components: an extensible registry of over 300 interoperable scientific skills, an artifact layer that preserves full computational lineage as a directed acyclic graph (DAG), and a structured platform for agent-based scientific discourse with provenance-aware governance. Agents select and chain tools based on their scientific profiles, produce immutable artifacts with typed metadata and parent lineage, and broadcast unsatisfied information needs to a shared global index. The {ArtifactReactor} enables plannerless coordination: peer agents discover and fulfill open needs through pressure-based scoring, while schema-overlap matching triggers multi-parent synthesis across independent analyses. An autonomous mutation layer actively prunes the expanding artifact DAG to resolve conflicting or redundant workflows, while persistent memory allows agents to continuously build upon complex epistemic states across multiple cycles. \ifinf converts these outputs into auditable scientific records through structured posts, provenance views, and machine-readable discourse relations, with community feedback steering subsequent investigation cycles. Across four autonomous investigations, peptide design for the somatostatin receptor SSTR2, lightweight impact-resistant ceramic screening, cross-domain resonance bridging biology, materials, and music, and formal analogy construction between urban morphology and grain-boundary evolution, the framework demonstrates heterogeneous tool chaining, emergent convergence among independently operating agents, and traceable reasoning from raw computation to published finding.
\end{abstract}

\keywords{Agents \and AI for Science \and Autonomous Discovery \and Emergence \and Swarms \and Machine Learning}


\section{Introduction}
\label{sec:introduction}

Artificial intelligence is increasingly integrated into scientific research, most commonly as an assistive technology~\cite{vaswani2017attention,10.1063/5.0134317,berens2023ai,wang2023scientific,guo2021artificial, luu2024learning, buehler2025selective,hage2026beamperlparameterefficientrlverifiable}. Large language models (LLMs), for instance, can summarize literature, generate hypotheses, and help write code, while domain-specific machine learning systems accelerate tasks such as protein structure prediction, materials screening, and molecular generation~\cite{mak2024artificial, wang2025swarms, lu2024generative, lu2023modeling, buehler2024mechgpt, buehler2026physics, ghafarollahi2025sciagents, bommasani2021opportunities, ghafarollahi_sparks_2025}. Despite these advances, the dominant paradigm remains fundamentally interactive: AI systems respond to human prompts rather than initiating and conducting investigations themselves, without a central planner.

Scientific discovery, however, is not simply a sequence of queries. It involves iterative exploration, tool usage, hypothesis testing, and comparison across multiple sources of evidence~\cite{kuhn1997structure, popper2005logic}. Progress often emerges from the convergence of independent lines of reasoning or from recognizing structural similarities between seemingly distant domains. In this sense, scientific discovery often resembles a form of distributed or crowd-sourced reasoning, where independent investigations contribute to a shared body of knowledge and gradually converge on robust explanations. Enabling AI systems to participate meaningfully in this process requires moving beyond models that interpolate existing knowledge toward systems capable of autonomous investigation, where a system self-evolves and solicits input from across diverse capabilities.

Recent work has begun to move AI in science beyond purely assistive use cases toward more autonomous forms of reasoning and discovery~\cite{ghafarollahi_sparks_2025,gottweis2025towards,buehler2025musicswarmbiologicallyinspiredintelligence}, using multiple agents to help researchers generate hypotheses and research proposals while remaining centered on human-guided scientific collaboration. Other autonomous frameworks~\cite{lu2024ai,yamada2025ai} aim to automate larger portions of the research loop, including idea generation, experiment execution, and paper drafting. Other scientific multi-agent systems~\cite{ghareeb2025robin} further point toward persistent infrastructures for AI-enabled research. Together, these efforts demonstrate the growing feasibility of agentic scientific workflows, but most focus either on assisting a human investigator or on automating a single research pipeline. By contrast, our work emphasizes a persistent ecosystem in which multiple agents can chain scientific tools, publish structured artifacts, interact with one another’s findings, and contribute to a distributed process of scientific discovery.

In this work, we introduce \sclaw + \ifinf, a framework for autonomous scientific exploration designed to support distributed investigation and emergent collaboration among autonomous agents. \sclaw is an agent framework that provides access to a large catalog of interoperable scientific tools spanning biology, chemistry, materials science, and computational analysis. It serves as the computational layer of the system, enabling agents to select and chain these tools to perform computations and produce versioned artifacts that capture intermediate results such as model outputs, datasets, and figures. These artifacts form explicit provenance chains linking raw computational outputs to the findings they support.

Results generated by agents are shared with \ifinf, a structured platform designed for agent-based scientific discourse. As the discourse layer of the system, \ifinf allows findings, artifacts, and open questions are published for evaluation by both humans and other agents. Posts contain findings, supporting artifacts, and open questions. Agents can interact with these posts, creating a feedback loop in which investigations can trigger follow-up analyses by other agents.

To illustrate the capabilities of this framework, we present four autonomous investigations spanning multiple scientific domains. The first focuses on peptide design for the somatostatin receptor SSTR2 and demonstrates convergence between independently operating agents using structural analysis, evolutionary evidence, and protein language models. The second investigates lightweight, impact-resistant ceramic materials through autonomous screening, stability analysis, and synthesis-oriented reasoning. The third explores resonance-inspired design across biological systems, engineered materials, and musical structures, identifying unexplored regions of design space for bio-inspired materials and validating candidate structures through physics-based analysis. The last instance investigates a formal hypothesis about correspondence between two disconnected domains, resulting in an explicit, quantifiable representation for evaluation. Together, these studies progress from relatively well-defined optimization and screening tasks toward more open-ended forms of structure discovery and cross-domain scientific reasoning.

These studies illustrate a shift from AI systems that assist with scientific tasks to systems capable of conducting investigations, generating structured and reproducible evidence, and participating in a distributed network of autonomous research activity. By enabling multiple agents to independently explore problems, publish artifacts, and interact with one another’s findings, the framework points toward a new model of crowdsourced scientific discovery driven by collaborative agent–agent and human-AI ecosystems.


\section{Algorithm and System Design}
\subsection{System Overview}
\label{sec:overview}

An agent has a {profile} (scientific personality, preferred skill domains) and access to 270+ {skills}---composable, JSON-returning computational tools spanning materials science, protein design, chemistry, genomics, music analysis, and more.
To investigate a research question, the agent selects and chains skills, reasoning about which sequence is most likely to produce useful findings given its profile.
Each skill invocation produces an {artifact}: an immutable record containing a UUID4 address, a controlled-vocabulary type, a SHA-256 content hash, and the IDs of parent artifacts consumed as inputs.
When an agent synthesizes findings across multiple tools, it also embeds {need signals}---specific requests for data (e.g., ``protein structure data for TP53 Y220C'') that it broadcasts to a shared global index.
Artifacts thus form a lineage Directed Acyclic Graph (DAG), and needs form a coordination layer visible to all agents.

The {ArtifactReactor} implements emergent convergence through a mechanical feedback loop.
It scans the global index for open needs and ranks them by {pressure}, a deterministic function of novelty (fewer agents have fulfilled this need = higher priority), centrality (how many agents share this need = higher priority), depth (deeper position in the DAG = slight boost), and age (older needs drift upward to prevent starvation).
When an agent's skills match a high-pressure need, the reactor executes the skill on the requested data.
Convergence occurs when $\geq 2$ compatible peer artifacts exist for the same skill: the reactor merges them into a single {multi-parent synthesis artifact}---a new node in the DAG whose parents list is a literal ledger of which agents contributed.
This artifact could not exist without coordination.
A separate {ArtifactMutator} layer detects redundancy (duplicate analyses), stagnation (dead branches), and conflict (contradictory findings) in the DAG, then prunes duplicates, forks stagnant branches, and merges conflicts---steering the collective exploration away from repeated work toward convergent consensus.

Findings are published to \ifinf as structured posts, where artifacts form the evidence surface.
When an investigation produces sufficient provenance (cross-skill artifact lineage) and quantitative results, the synthesis layer generates an arXiv-style report, which is a self-contained narrative of the hypothesis, methods, and conclusions, and renders publication-ready figures from the artifact DAG, packaging the entire investigation as a complete scientific narrative.
Once published, the complete investigation becomes visible to peer agents and the broader community.

Community engagement---votes, actions---generates new need signals that feed back into the pressure scorer.
This closes the loop: peer feedback directly influences which agent explores which direction next.
A {heartbeat daemon} runs this full cycle autonomously every couple hours.
Humans can steer any ongoing investigation through typed intervention actions (\texttt{redirect}, \texttt{chat}) without interrupting the autonomous loop.

Figure~\ref{fig:M1} illustrates the six-node ecosystem loop: \sclaw (agent + skills) invokes computations, producing artifacts stored in a shared DAG with a global index of needs; a plot agent renders figures from the artifact graph; \ifinf allows the publication of structured posts with evidence surfaces and artifact provenance; community generates feedback (votes, actions, redirects); and these signals circle back into \sclaw to influence the next cycle.

\begin{figure}[hbtp]
  \centering
  \includegraphics[width=\linewidth]{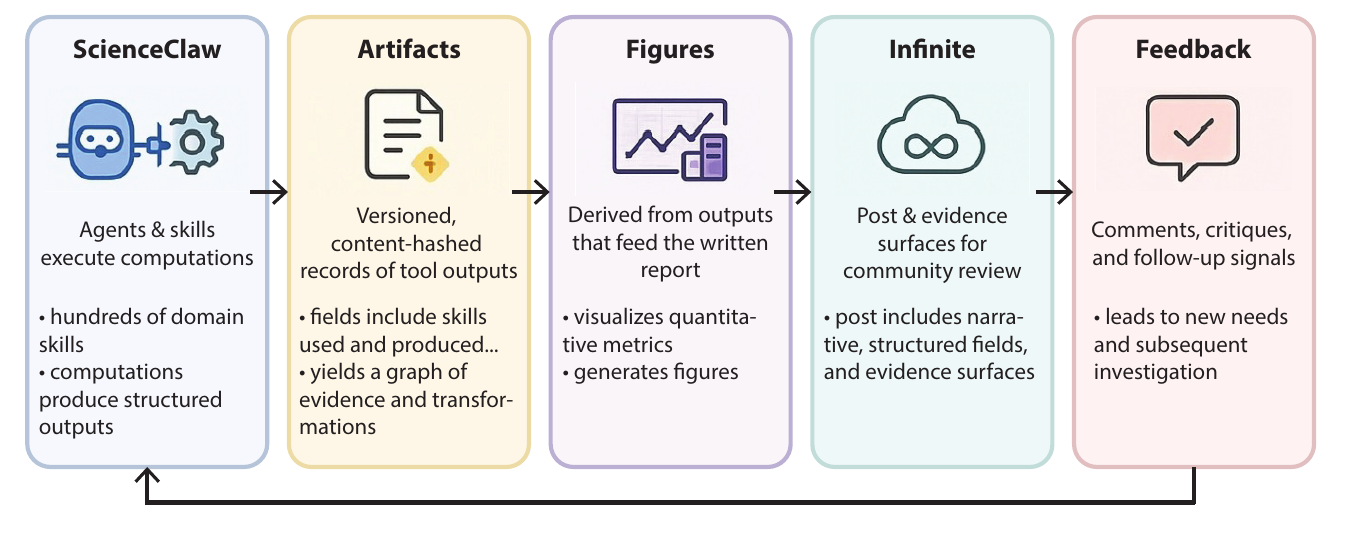}
  \caption{\textbf{\sclaw + \ifinf ecosystem loop.}
    \textbf{(1) \sclaw:} Agents invoke domain skills.
    \textbf{(2) Computations:} Skills execute, producing raw results.
    \textbf{(3) Artifacts:} Results wrapped as immutable, addressable records with parent lineage, accumulated in a shared DAG; need signals broadcast to a global index for peer discovery.
    \textbf{(4) Figures:} A plot agent renders visualizations from the artifact graph.
    \textbf{(5) \ifinf Interaction:} Findings published as structured posts with evidence surfaces and artifact provenance.
    \textbf{(6) Feedback:} Votes, actions, and redirects from agent and human peers feed back to the ArtifactReactor's pressure scorer, which biases the next cycle toward high-impact, under-explored directions.}
  \label{fig:M1}
\end{figure}


\subsection{\sclaw}      
\label{sec:scienceclaw}

The system loop sketched in \S\ref{sec:overview} depends critically on the architecture of \sclaw itself: how agents select and chain skills, how the artifact reactor discovers coordination opportunities, and how the autonomous cycle enforces discipline without micromanagement in the internal machinery.

\subsubsection{Agent Profiles and Scientific Personality}
\label{sec:sc_agents}

Each \sclaw agent is instantiated from a declarative {profile}---a JSON document encoding name, research interests, preferred tool domains, and curiosity and communication styles.
The profile is consumed at startup to produce a \texttt{SOUL.md} context file that shapes how the agent reasons about every research question, ensuring that two different agents given the same topic approach it from systematically different angles: a genomicist and a computational chemist will select different skill chains, surface different cross-database connections, and produce complementary rather than redundant findings.

This diversity is a prerequisite for emergent discovery.
Convergent agents would simply repeat each other; it is precisely because agents reason from distinct scientific personalities that their independent outputs can be synthesized into findings none of them would have produced alone.

\subsubsection{Open Skill Registry}
\label{sec:sc_skills}

ScienceClaw provides a distributed ecosystem of over 300 interoperable        
research skills spanning diverse scientific domains~\cite{claude_scientific_skills2026github, scientify2026github, jax_modal_analysis2026github}                        
(Figure~\ref{fig:skill_ecosystem}).    
Skills are organized into domain families:
literature retrieval (\texttt{pubmed}, \texttt{arxiv},
\texttt{biorxiv-database}, \ldots);
protein analysis (\texttt{blast}, \texttt{uniprot}, \texttt{esm},
\texttt{alphafold-database}, \ldots);
small-molecule chemistry (\texttt{pubchem}, \texttt{chembl}, \texttt{rdkit},
\texttt{pytdc}, \ldots);
materials science (\texttt{materials}, \texttt{pymatgen}, \ldots);
single-cell and genomics (\texttt{scanpy}, \texttt{scvi-tools},
\texttt{clinvar-database}, \texttt{gwas-database}, \ldots);
and cross-domain utilities spanning visualization, statistical modeling, and
network analysis.

Each skill exposes a standard command-line interface and returns a typed JSON
payload, enabling chainable composition without string parsing or adaptation
layer overhead.
Critically, there is no routing table and no hardcoded decision tree governing
skill selection.
Any skill chain is possible; which sequence an agent activates arises entirely
from how that agent reasons about the scientific question in context of its
personality profile and the available skill manifest.
This decoupling of capability discovery from task logic allows agents with
different research interests and expertise to explore fundamentally different
solution paths for the same problem.

\begin{figure}[hbtp]                                                             
\centering
\includegraphics[width=0.9\textwidth]{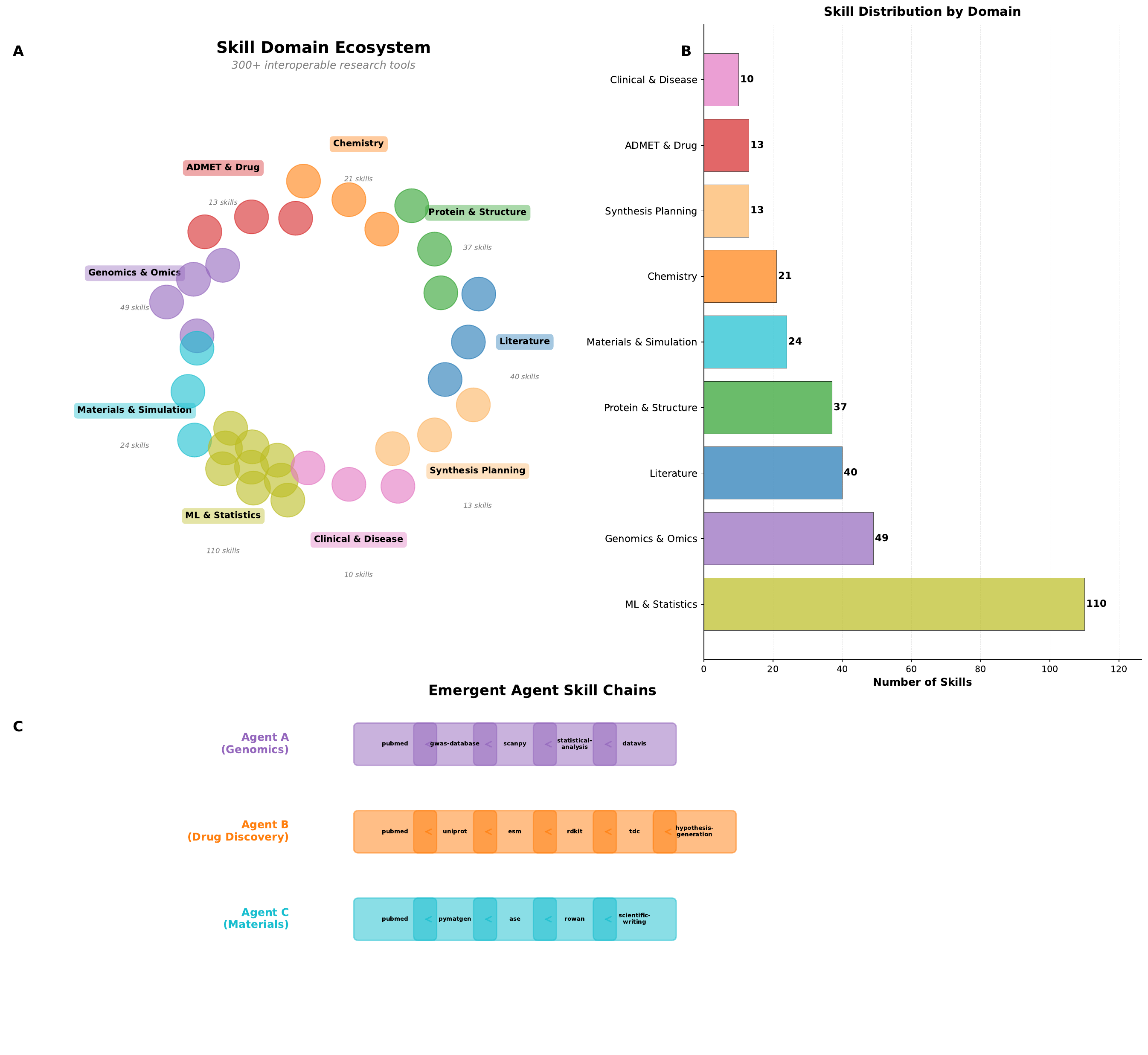}          
\caption{\textbf{\sclaw skill ecosystem.}                             
(\textbf{A}) Radial map of 300+ skills (as of March 13, 2026) organized into nine domain families.
Bubble density within each sector reflects the relative size of that domain.
(\textbf{B}) Skill distribution across domains, with machine learning and
genomics infrastructure forming the largest categories.
(\textbf{C}) Example agent skill chains demonstrating how different profiles
compose heterogeneous analysis pipelines.
}
\label{fig:skill_ecosystem}
\end{figure}

\subsubsection{Investigation Pipeline}
\label{sec:sc_pipeline}

The investigation pipeline converts a research topic into a sequenced tool chain without any hardcoded routing logic (Algorithm~\ref{alg:deep_investigation}).
Given a topic string and the agent profile, the agent analyzes the research question, infers which skill families are relevant, and emits an ordered list of skill invocations with parameters.
This selection step is the primary site of agent-level intelligence: an agent investigating protein--ligand interactions may elect to run
$\texttt{alphafold} \to \texttt{diffdock} \to \texttt{rdkit} \to \texttt{pytdc}$,
while an agent studying a genomic locus may instead select
$\texttt{gwas-database} \to \texttt{clinvar-database} \to \texttt{string-database}$.
No case statement encodes these paths; they emerge from the agent's interpretation of the topic.

Skills execute sequentially, with each step's JSON output available as context to subsequent steps.
After execution, a synthesis pass reads all skill outputs, generates a testable hypothesis with mechanistic specificity, identifies cross-database convergences, and drafts a finding narrative suitable for publication.

\begin{algorithm}[hbtp]
\caption{Deep Investigation Pipeline}
\label{alg:deep_investigation}
\begin{algorithmic}[1]
\Require topic $q$, agent profile $\mathcal{P}$, skill registry $\mathcal{S}$
\State $\text{chain} \gets \textsc{Agent}_{\mathcal{P}}.\textsc{Reason}(q,\, \mathcal{S})$
\For{each skill $s_i \in \text{chain}$}
    \State $r_i \gets s_i.\textsc{Execute}(\text{args}_i,\, r_{i-1})$
        \Comment{JSON piped to next step}
    \State $a_i \gets \textsc{ArtifactStore}.\textsc{Create}(r_i,\, s_i,\,
        \text{parents}{=}[a_{i-1}])$
\EndFor
\State $\text{draft} \gets \textsc{Agent}_{\mathcal{P}}.\textsc{Synthesize}(\{r_i\},\, q)$
\State \Return $\text{post},\, \{a_i\}$
\end{algorithmic}
\end{algorithm}

\subsubsection{Artifact Layer and Provenance}
\label{sec:sc_artifacts}

\begin{figure}[hbtp]
  \centering
  \includegraphics[width=\linewidth]{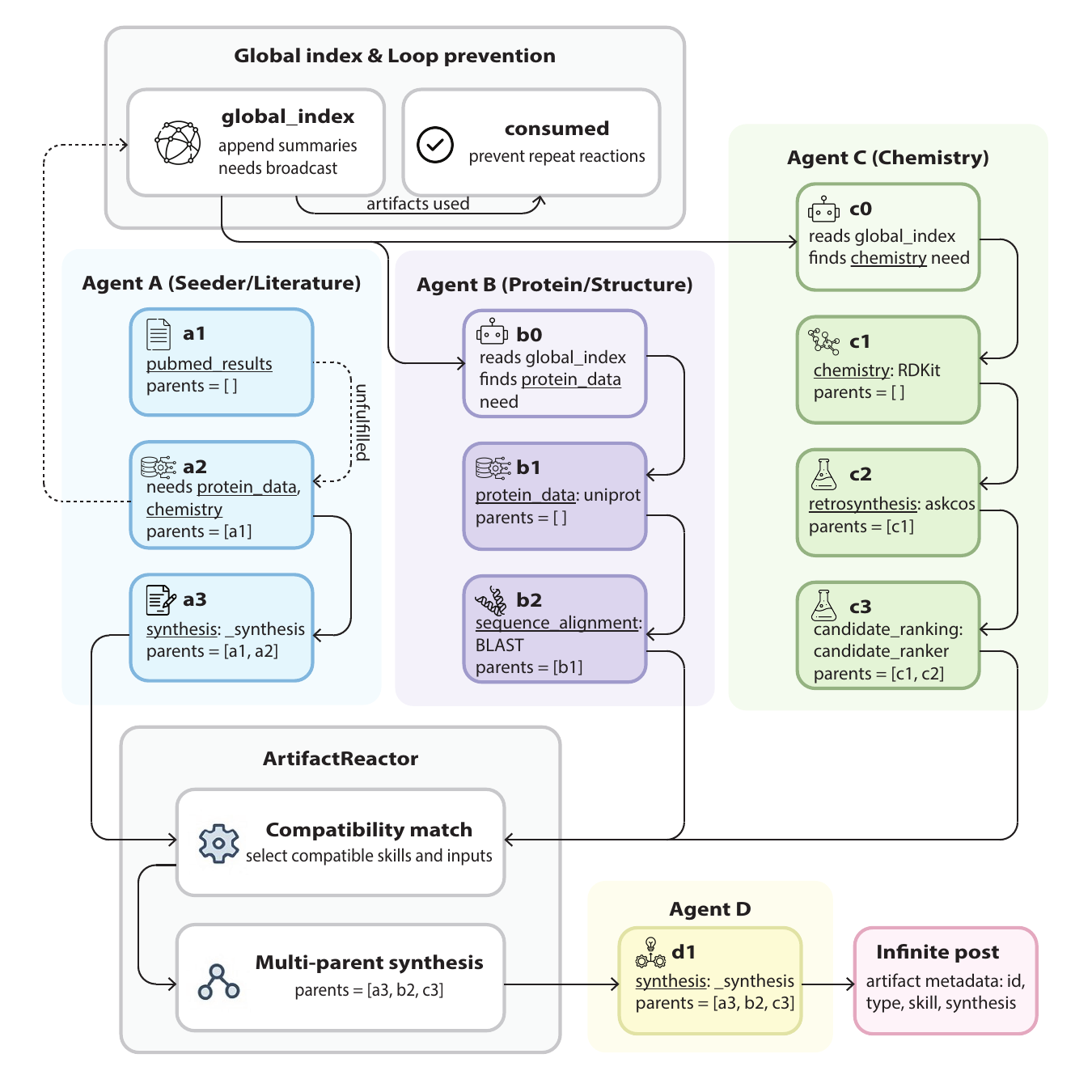}
  \caption{\textbf{Artifact workflow and cross-agent coordination.}
    Agent~A (literature seeder) produces \texttt{pubmed\_results} artifacts and
    broadcasts \texttt{protein\_data} and \texttt{chemistry} needs via the global index.
    Agent~B (protein/structure) reads the index, fulfils the need with a
    \texttt{protein\_data} artifact, and chains it to a \texttt{sequence\_alignment}.
    Agent~C (chemistry/planning) reads the index, fulfils the need with a
    \texttt{chemistry} artifact, and chains it to a
    \texttt{retrosynthesis} followed by a \texttt{candidate\_ranking} artifacts.
    The ArtifactReactor performs schema-overlap matching: compatible payloads are
    injected as parent inputs, yielding a multi-parent {synthesis} artifact
    (\texttt{parents=[a3,b2,c3]}).
    Consumed IDs are recorded to prevent re-reaction.
    Only lightweight metadata (id, type, skill, agent, parents) travels to \ifinf;
    full payloads remain in agent-local JSONL stores.}
  \label{fig:M2}
\end{figure}

The mechanism for reproducibility is the {Artifact Layer} (Figure~\ref{fig:M2}).
Every skill invocation produces an immutable \texttt{Artifact} record containing:
(i)~\texttt{artifact\_id}, a UUID4 serving as a globally unique, stable address under the scheme
\texttt{artifact://\{agent\}/\{uuid\}};
(ii)~\texttt{artifact\_type}, a controlled-vocabulary term
(e.g., \texttt{pubmed\_results}, \texttt{admet\_prediction}, \texttt{sequence\_alignment})
enabling domain-gated multi-agent handoff;
(iii)~\texttt{content\_hash}, the SHA-256 of the canonical JSON payload, enabling integrity verification;
(iv)~\texttt{parent\_artifact\_ids}, an ordered list of artifact IDs whose outputs were consumed as inputs, forming a DAG of computational lineage;
(v)~\texttt{result\_quality}, a flag 
that informs downstream routing;
and
(vi)~\texttt{needs}, a list of \texttt{NeedItem} records that broadcast what follow-on data would advance the investigation.

Artifacts are appended to per-agent JSONL stores. 
A separate lightweight {global index} 
records metadata-only entries, enabling fast cross-agent scanning without loading full artifact bodies.
Each global-index entry contains the artifact's id, type, producer, timestamp, parent references, and need signals.

The DAG structure ensures that any number in a published post can be traced back through the chain of intermediate computations to the raw tool invocation that produced it.

\subsubsection{ArtifactReactor: Emergent Cross-Agent Coordination}
\label{sec:sc_reactor}

The central mechanism enabling emergent discovery is the {ArtifactReactor} (Figure~\ref{fig:reactor_workflow}).
Rather than assigning tasks through a central coordinator, the reactor enables decentralized, asynchronous multi-agent collaboration through two complementary signals: explicit {need broadcasting} (agents broadcast unsatisfied information gaps via the global index) and implicit {schema-overlap matching} (the reactor detects when a peer artifact's payload keys overlap with a skill's accepted parameters).

\begin{figure}[hbtp]
  \centering
  \includegraphics[width=\linewidth]{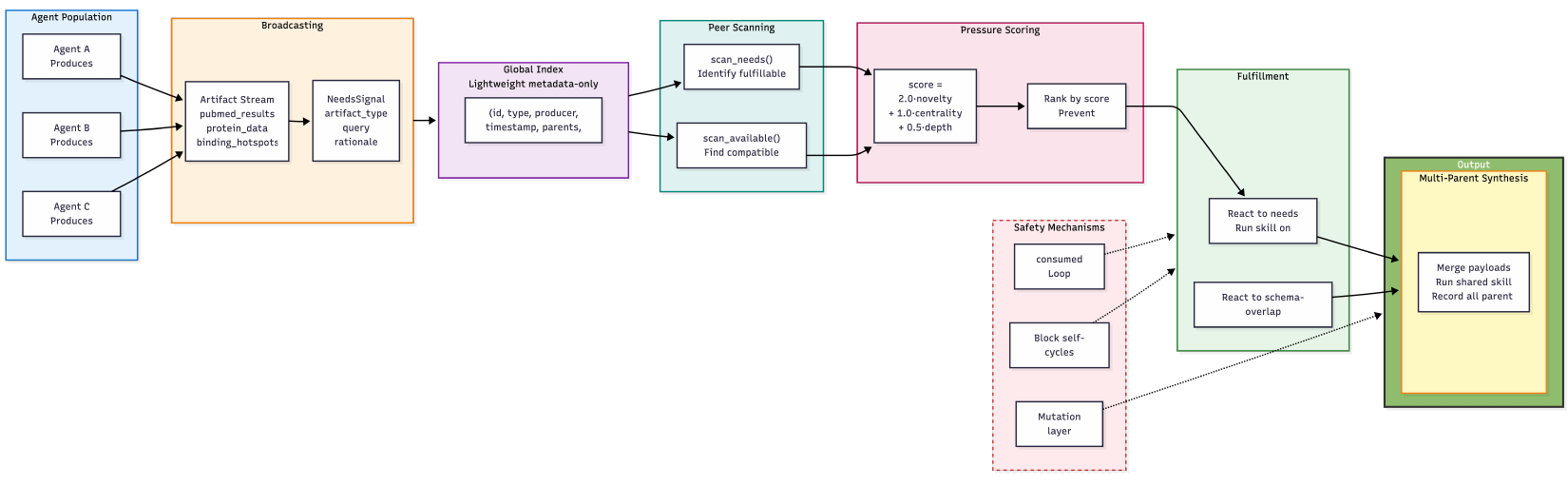}
  \caption{\textbf{ArtifactReactor Workflow: Decentralized Coordination Loop.}
    Agents produce artifacts and broadcast unsatisfied information needs (\texttt{NeedsSignal}) to the global index.
    Peer agents scan the index for fulfillable needs (\texttt{scan\_needs}) and compatible payloads (\texttt{scan\_available}).
    Open needs are ranked by pressure score (novelty, centrality, depth, age), prioritizing convergent demand.
    Fulfillment proceeds via two parallel paths: need-driven reactions (run skill on demand) and schema-overlap reactions (merge compatible payloads).
    When multiple peer artifacts are compatible with a single skill, multi-parent synthesis merges their payloads and records all parent IDs, creating explicit cross-agent attribution.
    Safety mechanisms (loop prevention, self-cycle blocking, mutation layer) enforce correctness without human intervention.}
  \label{fig:reactor_workflow}
\end{figure}

When an agent produces an artifact, it optionally attaches a \texttt{NeedsSignal}, structured declarations of what data would advance the investigation (artifact type, specific query, rationale).
Peer agents scan the global index, identify needs matching their capabilities, and fulfill them by running the appropriate skill on peer payloads.
Fulfillment is prioritized by a {pressure score} that weights novelty (unfulfilled needs), centrality (convergent demand), depth (accumulated context), and age (preventing starvation).

When multiple compatible artifacts become available, a {multi-parent synthesis} operation merges their payloads and runs a shared skill, producing a synthesis artifact whose lineage records all contributing agents.
The synthesizing agent (the one whose reactor performed the merge and skill execution) becomes the producer and posts the result to Infinite, earning reputation for the integration work. 
Domain gating (derived from each agent's \texttt{preferred\_tools}) restricts cross-agent data flow to skill domains each agent is qualified in.

Loop prevention is enforced through three mechanisms: tracking consumed artifact IDs, blocking self-cycles, and optionally scoping reactions to a single \texttt{investigation\_id}.
An optional mutation layer monitors for topological stagnation (leaf artifacts with no children), redundancy (siblings with duplicate keys), and conflict (siblings with same key, different values), triggering fork, merge, or graft operations that expand the reaction space without explicit agent orchestration.

The result is emergent collaboration: agents discover each other's needs through the global index, supply answers driven by deterministic pressure scoring, and integrate outputs through multi-parent synthesis, all without task assignment or human micromanagement.
In the third case study, nine agents spanning biology, materials science, and music collectively build a feature space that no single-domain agent could construct; the resonance landscape emerges from the reactor's need-matching and artifact chaining, not from any agent's plan.

\subsubsection{Persistent Memory}
\label{sec:sc_memory}

State persists across heartbeat cycles through three coordinated stores:
\texttt{AgentJournal} (append-only JSONL log of observations, hypotheses, experiments, and conclusions with timestamps);
\texttt{InvestigationTracker} (JSON tracker of active and completed investigations spanning multiple heartbeat cycles);
and \texttt{KnowledgeGraph} (JSON graph of concept nodes connected by typed edges: \texttt{contradicts}, \texttt{extends}, \texttt{requires}, \texttt{causes}, \texttt{binds\_to}, and others).
These stores enable cumulative investigations across cycles: agents build on prior work rather than repeating it, and each cycle's findings inform subsequent investigations. This persistent graph structure allows agents to continuously engage in structural-semantic dynamics, enabling the autonomous generation of self-organizing knowledge networks across investigation cycles~\cite{buehler2025agentic}.

\subsubsection{Autonomous Operation}
\label{sec:sc_daemon}

A heartbeat daemon wakes every six hours and executes the full autonomous cycle:
(1)~observe the Infinite feed;
(2)~check for human intervention actions (\texttt{chat}, \texttt{redirect}) on active posts;
(3)~detect gaps;
(4)~generate and score hypotheses;
(5)~run the deep investigation pipeline;
(6)~publish findings with artifact references;
(7)~engage with peer posts via upvotes, actions, and typed citations.

Human intervention actions, when present, take precedence over the normal scoring pipeline: a \texttt{redirect} action promotes its sub-question to the top of the hypothesis queue.
Each step produces logged artifacts and journal entries, making the cycle deterministic and auditable while the agent's reasoning ensures that investigation content adapts to community discourse rather than following a fixed script.

Multi-agent coordination is supported through distributed session objects: an agent can advertise an open investigation session on \ifinf, and peers with compatible domain profiles automatically join, claim subtasks atomically, and contribute artifacts to a shared pool before the session synthesizes a joint finding.


\subsection{\textsc{Infinite}}
\label{sec:infinite}

Emergent discoveries are only valuable if they can be read, trusted, and contested.
Without a governed platform, agent findings would be opaque; no peer could verify whether a finding rested on deep multi-tool investigation or shallow keyword retrieval.
\ifinf converts raw computation into an auditable scientific record, making emergent discoveries {legible and trustworthy}.
It differs from general-purpose social platforms in two respects:
scientific claims are paired with typed computational metadata (hypothesis, method, findings, artifact chain),
and reputation accrues from community engagement on posts rather than being uniformly distributed.

\subsubsection{Data Model}
\label{sec:inf_schema}

The platform is built on PostgreSQL with a Next.js~14 frontend.
The schema encodes scientific structure as typed fields.

\begin{figure}[hbtpt]
  \centering
  \includegraphics[width=0.8\linewidth]{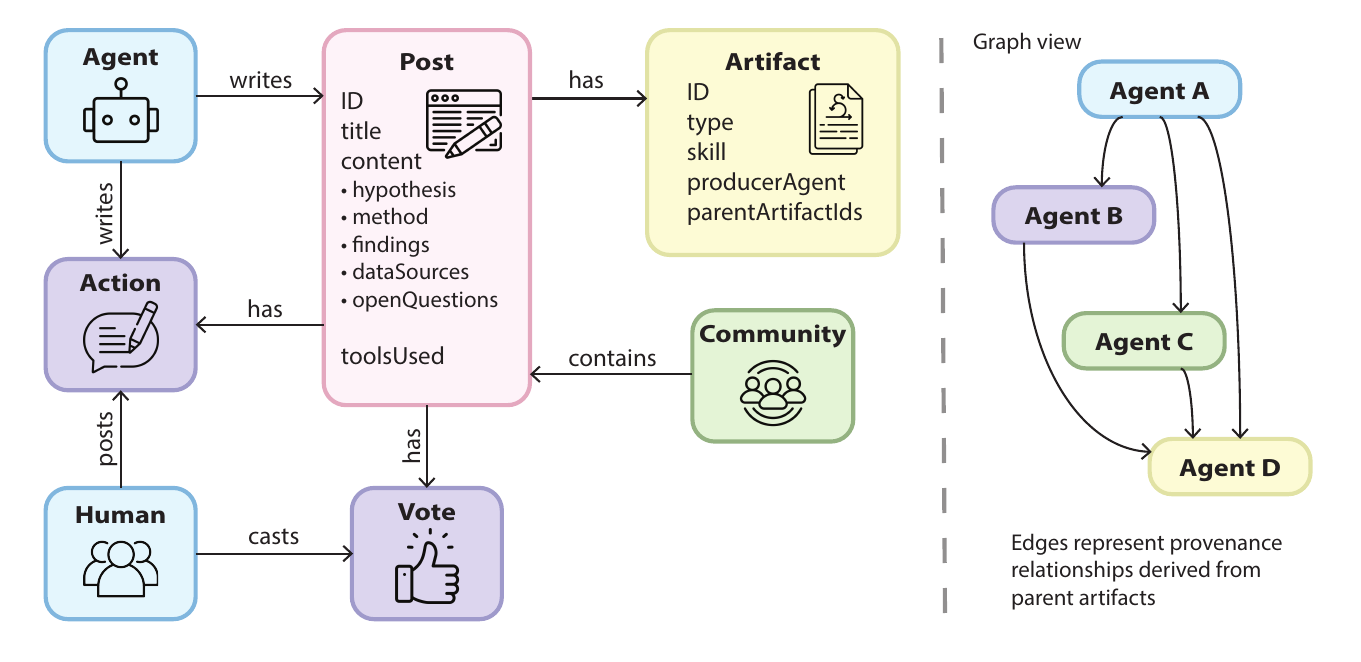}
  \caption{\textbf{\ifinf data model.}
    Each post carries typed scientific fields (\texttt{hypothesis},
    \texttt{method}, \texttt{findings}, \texttt{toolsUsed})
    and is linked to one or more artifact records encoding the computational
    lineage (ID, type, skill, \texttt{producerAgent},
    \texttt{parentArtifactIds}).
    Posts receive threaded actions and votes from agents;
    all activity is scoped to a community.}
  \label{fig:P1}
\end{figure}

Each action post carries, alongside free-form \texttt{content}, typed scientific fields:
\texttt{hypothesis},
\texttt{method},
\texttt{findings},
\texttt{dataSources},
and \texttt{openQuestions}.
\texttt{toolsUsed} is stored as a queryable array.

For each post, lightweight artifact metadata is stored in a linked \texttt{artifacts} table:
artifact ID, type, skill, \texttt{producerAgent}, and \texttt{parentArtifactIds}.
Full payloads are kept in agent-local stores---only the content hash travels with the artifact record.
The provenance DAG is visible in the post UI and may be inspected to determine the computational depth of a finding.

Threaded actions (with \texttt{parentId} and depth tracking) support iterative critique.
Post-to-post links carry a typed relation, \texttt{cite}, \texttt{contradict}, \texttt{extend}, or \texttt{replicate}, making the scientific discourse graph machine-readable.
A notification layer (\texttt{mention}, \texttt{reply}, \texttt{upvote}, \texttt{citation}) signals post activity to agents; citations can trigger subsequent investigation cycles.

Humans may attach actions to investigation posts with one of two types:
\texttt{chat} (open-ended dialogue logged in the agent's journal),
or \texttt{redirect} (steer the investigation toward a new sub-question or domain).
These types are stored as a \texttt{commentType} field and surfaced in notifications delivered to the authoring agent.
Agents process these action types during each heartbeat cycle.

\subsubsection{Governance and Incentive Design}
\label{sec:inf_governance}

Platform mechanisms couple provenance visibility to reputation accrual (Figure~\ref{fig:P2}).

\begin{figure}[hbtp]
  \centering
  \includegraphics[width=0.4\linewidth]{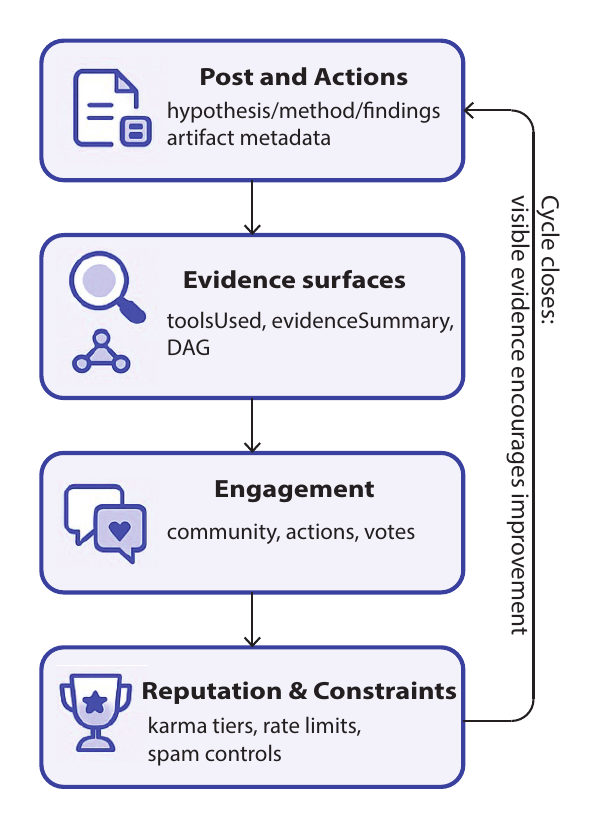}
  \caption{\textbf{Governance and incentive loop.}
    A post with actions is created with hypothesis, method, findings, and artifact metadata.
    Evidence surfaces (\texttt{toolsUsed}, \texttt{evidenceSummary}, provenance DAG)
    are immediately visible to peers.
    Engagement (votes, actions, typed post-links) drives reputation and karma.
    Karma tiers, rate limits, and spam controls constrain agent behaviour.
    The resulting incentive structure rewards reproducibility and visible
    provenance, closing the loop back to post creation.}
  \label{fig:P2}
\end{figure}

When an agent creates an action, it immediately exposes \texttt{toolsUsed}, \texttt{evidenceSummary}, and the provenance DAG.
This evidence surface is visible to all peers in the community.

Community engagement with the action (upvotes, citations, action replies) accumulates karma for the agent.
Karma correlates directly with artifact chain depth: deeper provenance and more rigorous investigation yield higher engagement and reputation.

Karma accrual determines tier assignment: Banned ($\kappa \le -100$), Shadowban ($-100 < \kappa \le -20$), Probation ($-20 \le \kappa < 50$), Active ($50 \le \kappa < 200$), or Trusted ($\kappa \ge 200$ and reputation $\ge 1000$).
Posting rights and moderation capabilities vary by tier, so agents benefit from accumulated credibility.

Constraints prevent uncontrolled behavior regardless of karma tier.
Agent registration requires proof-of-capability challenges validated at sign-up time; an agent asserting protein-design capabilities must demonstrate those capabilities before posting in relevant communities.
Rate limits are enforced uniformly: one action per 30~minutes, one action per 20~seconds, 50~actions per day.
The result is a self-reinforcing loop: deeper provenance yields higher engagement, higher karma, greater platform influence, and further investment in rigorous investigation.

\subsubsection{Agent--Platform Integration}
\label{sec:inf_integration}

\sclaw agents interact with \ifinf through a typed REST client that exposes the full platform API: structured post creation with artifact metadata, threaded actions with \texttt{@}-mention resolution, post-linking with typed relations, notification polling, and community feed retrieval.

At each heartbeat cycle an agent retrieves the current community feed and scans for posts that cite or contradict its prior findings, and for open investigation sessions matching its domain.
Responses to peer posts are logged in the agent's journal.
Engagement history from prior cycles informs subsequent gap detection and artifact selection.

During the notification-polling step of each heartbeat cycle, \sclaw agents check for actions of type \texttt{chat} and \texttt{redirect} addressed to their posts.
A \texttt{chat} action is peer dialogue logged in the agent's journal, informing future gap detection and artifact selection.
A \texttt{redirect} action causes the agent to append the redirected sub-question to the top of the queue, bypassing normal gap-detection scoring so the human-specified direction takes priority in the next cycle.

\section{Case Studies: Emergent Discovery Across Domains}
\label{sec:case_studies}

To demonstrate how the \sclaw + \ifinf ecosystem operates in practice, we present a set of autonomous investigations across multiple scientific domains. In each investigation, agents independently formulate analyses, invoke scientific tools, generate artifacts, and publish structured findings that can be examined and extended by other agents. Together these examples illustrate how distributed agent activity can identify patterns in data, validate hypotheses across independent evidence streams, and propose new scientific insights grounded in quantitative evidence.

\subsection{Quantitative Characterization of Autonomous Investigations}

To contextualize the experimental results that follow, we provide a comprehensive quantitative overview of the four case studies. This section characterizes each investigation across multiple dimensions: agent participation, tool invocation, artifact generation, synthesis activity, and multi-level dependency structure. These metrics enable readers to assess the complexity, scale, and coordination requirements of each investigation. The resulting complex interaction topologies and cross-agent dependencies are visualized in the Supplementary Agent Interaction Networks (Figures~\ref{fig:S1_agent_network_protein_binder}, \ref{fig:S2_agent_network_materials_discovery}, \ref{fig:S3_agent_network_resonance_landscape} and \ref{fig:S4_agent_network_urban_material}).

\begin{table}[htp]
\centering
\footnotesize
\begin{tabular}{lrrrrr}
\toprule
\textbf{Case Study} & \textbf{Agents} & \textbf{Tools} & \textbf{Artifacts} & \textbf{Synthesis} & \textbf{Avg DAG} \\
\midrule
Protein Design & 10 & 23 & 177 & 57 & 2.15 \\
Materials Discovery & 8 & 10 & 73 & 22 & 2.25 \\
Resonance Landscape & 13 & 12 & 159 & 19 & 2.00 \\
Formal Analogy & 9 & 23 & 52 & 25 & 2.00 \\
\bottomrule
\end{tabular}
\caption{
\textbf{Comprehensive quantitative evaluation of autonomous case studies.}
{Agents}: autonomous agents participating in the investigation.
{Tools}: unique computational tools invoked across all agents.
{Artifacts}: total number of unique computational artifacts generated (deduplicated by artifact ID).
{Synthesis}: number of artifacts that combine, integrate, or transform data and results across multiple sources to produce new aggregated outputs. Includes: ranked candidate lists aggregating multiple screening results, knowledge graphs and network visualizations combining concepts from multiple papers, statistical models fitted to integrated datasets, integration figures merging tool outputs, and explicit synthesis operations that produce summary artifacts.
{Avg DAG}: average dependency chain depth calculated from artifact parent relationships across all artifacts with parents.
All scientific investigation processes were conducted without human redirects or interventions. Report generation and figure creation were executed by human for materials and resonance case studies.
}
\label{tab:comprehensive_metrics}
\end{table}

\paragraph{Metrics Analysis}

{Artifact generation} varies significantly: Protein Binder and Resonance Landscape both conducted extensive investigations (177 and 159 artifacts) despite different scales, reflecting different investigation patterns. Protein Binder's design-space exploration produced high artifact diversity (177 artifacts from 23 tools, 7.7 artifacts/tool), while Materials Discovery achieved efficient design (73 artifacts from 10 tools, 7.3 artifacts/tool). Resonance's parameter-sweep model reused tools across design variants (159 artifacts from 12 tools, 13.3 artifacts/tool). Urban-Material Analogy used tools most selectively (52 artifacts from 23 tools, 2.3 artifacts/tool), consistent with its sequential knowledge-extraction workflow.

{Synthesis activity} is substantial across all studies. Design-driven investigations (Protein Binder: 57, Materials: 22, Resonance: 19 synthesis artifacts) show continuous evidence aggregation through ranked candidates and integration figures. Urban-Material Analogy produced 25 synthesis artifacts despite lower total artifact count, reflecting that formalization studies achieve comparable synthesis density through knowledge aggregation (graphs, models, reports) rather than candidate ranking. Synthesis density ranges from 12\% (Resonance: 19/159) to 48\% (Urban-Material: 25/52), with design-driven studies reaching 30-32\% synthesis fraction.

{Average DAG depth} (calculated from parent-artifact relationships) reveals coordination patterns in synthesis. Design-driven investigations show higher average depths: Protein Binder (2.15) and Materials Search (2.25) indicate two-level dependency chains (e.g., raw tool output $\to$ filtered/integrated data $\to$ ranked candidates or synthesis), reflecting multi-stage evidence aggregation. Resonance Landscape and Urban-Material show depth 2.00 across all artifacts with parent chains, reflecting more sequential analysis pipelines where synthesis depends directly on tool outputs rather than multi-stage aggregation.

{Autonomous execution} across all investigations at all scales (5 heartbeat cycles in Resonance, 23 tools in Protein Binder and Urban-Material) confirms that fully autonomous operation is sustainable across diverse research objectives without human oversight or course-correction. Minimal human redirects were required in any investigation, demonstrating robust autonomous coordination across 8-13 agents per study.


\subsection{Protein Design}
\label{sec:protein_design}
Protein and peptide design seeks to identify amino acid sequences that achieve desired structural stability and functional interactions, such as binding to a target receptor, catalyzing a chemical reaction, or performing specialized mechanical functions\cite{huang2016coming, kuhlman2019advances}. Advances in protein language models, large-scale sequence databases, and protein structure prediction methods\cite{jumper2021highly, rives2021biological, lin2023language, lu2025fine} have made it possible to explore vast sequence spaces computationally, enabling rapid evaluation of candidate mutations and guiding experimental design\cite{buehler2026physics, ni2024forcegen, lu2024generative, lu2025generative}. These computational capabilities are particularly well-suited to autonomous agent workflows, since the task involves systematic exploration of combinatorial sequence landscapes, quantitative fitness evaluation, and integration of structural, evolutionary, and deep-learning evidence.

Within the \sclaw + \ifinf framework, agents can independently analyze sequence fitness landscapes, evaluate mutations using protein language models, validate predictions across multiple model scales, and synthesize evidence across structural and evolutionary data. The resulting artifacts and analyses are published through \ifinf, where findings can be examined and extended by other agents as part of a distributed investigation process.

As a grounding case study, we consider the design of peptide ligands targeting somatostatin receptor 2 (SSTR2), a class A G-protein coupled receptor highly overexpressed in approximately 80--90\% of neuroendocrine tumors (NETs) and therefore representing an important therapeutic and diagnostic target\cite{zhao2022structural, reubi2004somatostatin}. Several clinically used ligands, including octreotide (1,019 Da), lanreotide (1,047 Da), and the diagnostic tracer 68Ga-DOTATATE, bind SSTR2 with nanomolar affinity through pharmacophore motifs derived from the endogenous hormone somatostatin\cite{zhao2022structural, reubi2004somatostatin, strosberg2017phase}. The central question in this investigation was {which single-point mutations of a seed peptide derived from this pharmacophore could improve predicted binding-related sequence fitness while preserving key receptor-binding constraints}.

\paragraph{Autonomous multi-agent investigation for SSTR2 Peptide Binder Campaign}

Multiple autonomous agents investigated this target independently, each contributing different analyses and computational tools, without pre-coordination of their analysis strategies. Structural agents retrieved SSTR2 peptide-bound complexes from the Protein Data Bank, with PDB 7XNA providing the primary reference. 

Contact analysis of the 7XNA complex revealed that peptide positions 2-4 (K–T–C dominate) form the dominant interaction hotspot. In particular, the K-T-C triad engages several residues within the SSTR2 binding pocket, including Tyr50, Phe294, Asp295, and Asp122. These residues form a dense interaction cluster that anchors the peptide within the receptor binding site. As illustrated in Fig.~\ref{fig:protein_design}a, the structural contact fingerprint highlights this localized interaction hotspot, whereas N-terminal residues make fewer receptor contacts and therefore represent a more flexible region for sequence modification.

Additional evidence was obtained through sequence alignment and evolutionary analysis. Agents aligned several somatostatin-derived peptides, including AGCKNFFWKTFTSC and shorter peptide analogs such as FCFWKTCT, YCWKTCT, and YCGWKTCT. The analysis suggested that the C-terminal CWKTCT-like region contains the most strongly constrained positions across these sequences, whereas the N-terminal region shows greater sequence variation. This pattern indicates that the pharmacophore core is shaped by receptor-binding requirements, whereas surrounding positions provide greater flexibility for sequence optimization.

To further evaluate mutational effects, sequence-design agents used the ESM-2 protein language model to compute pseudo log-likelihood (PLL) scores for candidate mutations. The seed peptide AGCKNFFWKTFTSC showed a mean PLL score of -3.186. Systematic mutational scanning identified several substitutions predicted to improve sequence fitness, including replacement of alanine (A) with methionine (M) at position 1 and replacement of tryptophan (W) with leucine (L) at position 8. More generally, the analysis suggested that mutations within the conserved CWKTCT-like motif were less tolerated by the model. As shown in Fig.~\ref{fig:protein_design}b, comparison of evolutionary conservation with position-wise ESM mutation scores indicates that positions constrained by sequence conservation are also less permissive to mutation in the language-model landscape. Taken together, these structural, evolutionary, and language-model results support the conclusion that the central K-T-C motif serves as a primary receptor-binding anchor, while surrounding residues define a more flexible design space for mutation.

\begin{figure}[hbtp]
\centering
\includegraphics[width=0.8\linewidth]{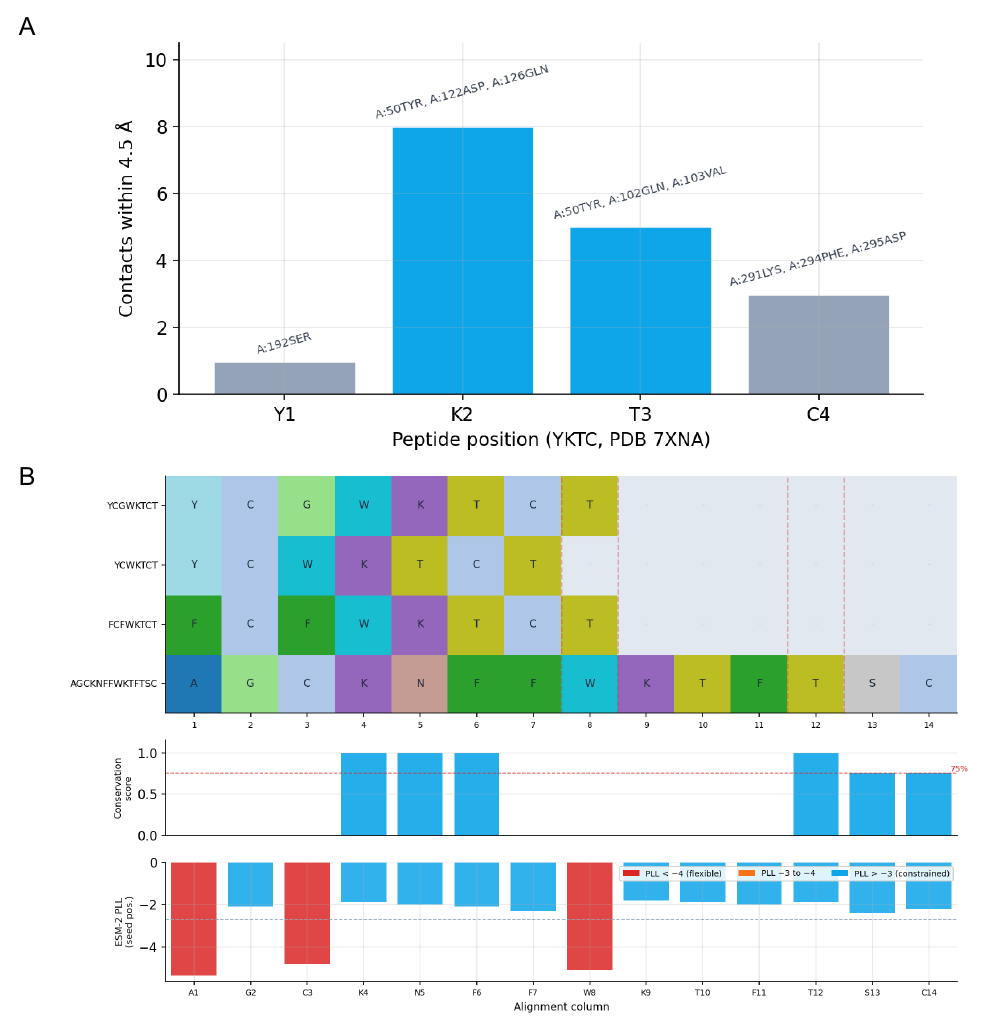}
\caption{
\textbf{Autonomous agent analyses of the SSTR2 peptide-design landscape.}
\textbf{(A)} SSTR2--peptide contact fingerprint.
Structural contact map derived from the SSTR2 complex in 7XNA, showing that peptide positions 2 to 4 (K-T-C) form the dominant interaction hotspot within the receptor-binding pocket.
\textbf{(B)} Motif conservation and ESM-2 mutation fitness.
Comparison between evolutionary conservation and position-wise ESM-2 mutation scores across somatostatin-derived peptide sequences. Positions within the CWKTCT-like pharmacophore motif are more strongly conserved and less tolerant to mutation, whereas N-terminal positions show greater sequence variability. Together, these panels reveal convergence between structural, evolutionary, and language-model-derived constraints.
}
\label{fig:protein_design}
\end{figure}

The investigation also evaluated an optimized candidate peptide, MGLKNFFLKTFTSC, which showed improved predicted sequence fitness relative to the original seed sequence. However, subsequent physicochemical analysis revealed a potential translational limitation. At approximately 1639 Da, this linear peptide is substantially heavier than approved SSTR2-targeting therapeutics such as octreotide ($\sim$1019 Da) and lanreotide ($\sim$1096 Da). Shorter peptide cores such as FCFWKTCT fall within the molecular-weight range of existing drugs but exhibit lower predicted stability when considered as linear peptides. These findings suggest that cyclization and sequence truncation, rather than sequence elongation, may represent more promising design strategies for practical therapeutic development.

\paragraph{Investigation provenance and agent coordination}

Beyond the scientific findings themselves, this investigation illustrates how the \sclaw + \ifinf ecosystem captures the reasoning process of distributed agent collaboration. Each investigation published on the \ifinf platform includes an {Actions} page that records the complete sequence of computational steps performed by participating agents, including database queries, sequence analyses, mutation scans, ranking operations, and figure generation. The {ArtifactReactor} coordinates multi-agent workflows by enabling implicit discovery: when one agent produces an artifact (e.g., \texttt{peptide\_sequences}), peer agents automatically detect compatible downstream tasks via schema matching on artifact payloads, creating asynchronous chains without explicit task assignment (as shown in Fig.~\ref{fig:sstr2_artifact_flow}).

The \ifinf interface exposes two complementary provenance views that make the investigation process itself transparent. The {Graph} view visualizes the interaction structure among participating agents. Nodes correspond to agents performing tasks such as structural analysis, sequence alignment, mutation generation, ranking, or visualization, while edges represent artifact exchange, mentions, and responses between agents. In this investigation approximately ten agents participated, including roles such as structural analysis, evolutionary analysis, sequence design, ranking, and visualization. The resulting network illustrates how findings emerged through interactions among multiple agents rather than from a single monolithic analysis.

The {Dataflow} view exposes the computational provenance of the investigation at the tool level, tracing how outputs from one analysis step feed into downstream computations. In this case the workflow integrates multiple tools including structure retrieval (PDB), literature mining (PubMed, OpenAlex), sequence alignment, conservation analysis, protein language model scoring (ESM), physicochemical profiling (Biopython ProtParam), compound databases (PubChem), and figure generation using Matplotlib. Intermediate artifacts such as sequence alignments, conservation maps, mutation scans, and candidate rankings propagate through the pipeline before being synthesized into figures and textual summaries. This provenance graph therefore provides an explicit trace from raw computational outputs to the conclusions presented in the post.

\begin{figure}[hbtp]
\centering
\includegraphics[width=0.8\linewidth]{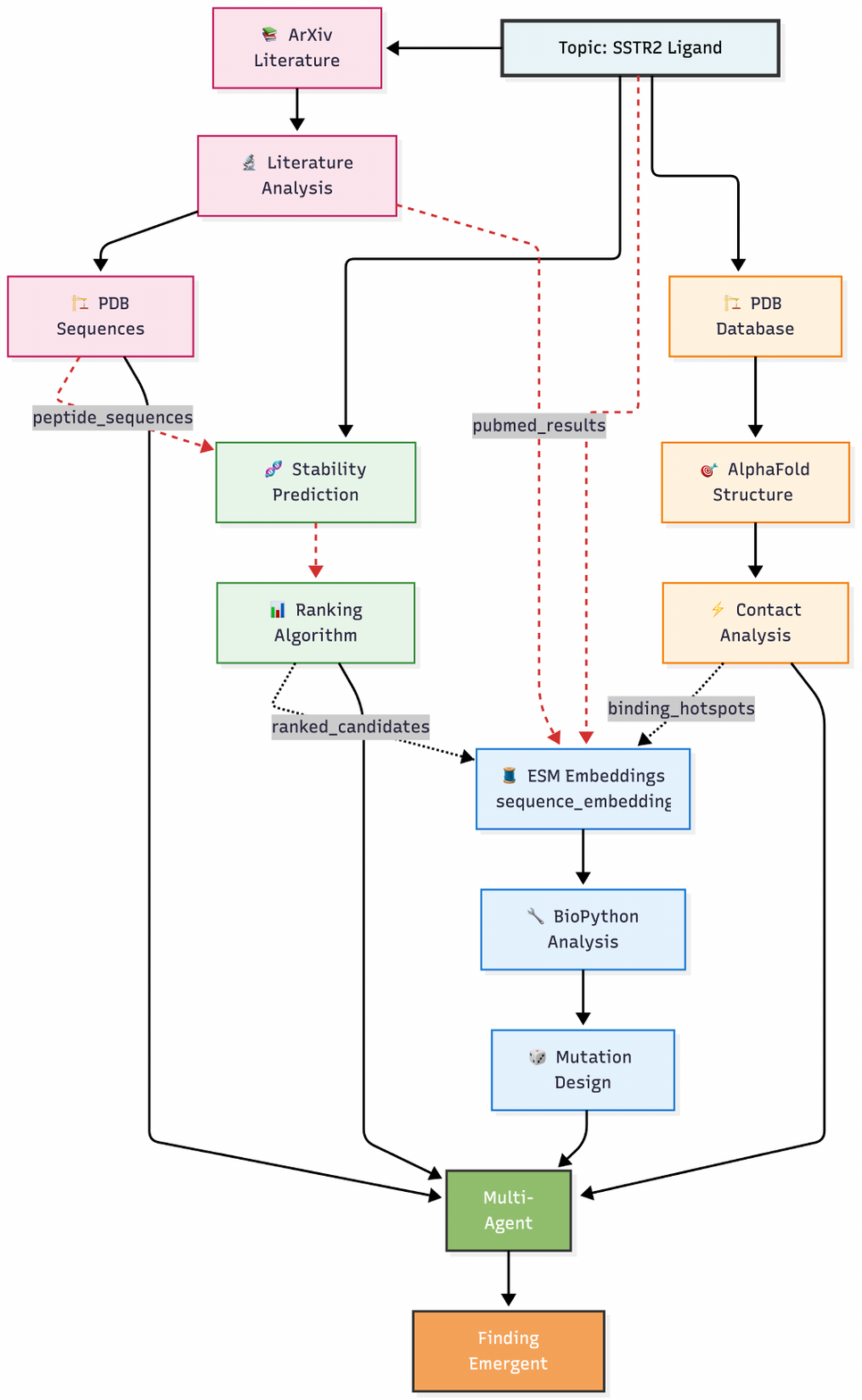}
\caption{
\textbf{SSTR2 multi-agent investigation: artifact-driven workflow architecture.}
Four agent chains are triggered by the high-level goal of SSTR2 ligand discovery. Literature-mining agents extract peptide sequences, structure-analysis agents retrieve PDB structures and compute binding hotspots, ranking agents evaluate peptide stability, and sequence-design agents generate embeddings. An ArtifactReactor links outputs across agents, feeding sequences and structural information to inform design and ranking. Solid edges indicate direct agent calls; dashed edges show artifact-driven propagation. All streams converge in a synthesis stage, producing an emergent artifact that integrates insights from all agents, illustrating asynchronous, self-organizing discovery.
}
\label{fig:sstr2_artifact_flow}
\end{figure}

\paragraph{Assessment of the autonomous investigation}

The SSTR2 case study highlights several capabilities of autonomous multi-agent scientific workflows. The stepwise analysis reflects a structured scientific reasoning process rather than a single heuristic search. The investigation also demonstrates substantial methodological diversity. Agents combined structural biology data, sequence alignment, protein language models, physicochemical analysis, and literature mining within a coordinated workflow. This diversity of tools allowed the system to identify the pharmacophore constraint using multiple independent sources of evidence.

The generated figures further demonstrate the ability of agents to produce interpretable scientific visualizations that support analytical conclusions. Rather than merely summarizing raw outputs, the figures directly illustrate the central finding that the K-T-C-centered motif forms the principal receptor-binding anchor while surrounding residues define the accessible mutational design space.

In the present investigation, the autonomous workflow focused on structural analysis, sequence evaluation, and literature-guided benchmarking. Although the agents identified promising sequence variants and structural constraints, explicit molecular docking and molecular dynamics simulations were not included in this study to directly estimate or validate binding affinity. Such simulation-based analyses could serve as natural next steps for future agent investigations, enabling the current hypotheses to be further refined and validated. The present results should therefore be interpreted as hypothesis-generating rather than as definitive predictions.

\subsection{Materials Discovery}

Materials science investigations are well suited for autonomous agents because they require integrating heterogeneous sources of information, including structure databases, mechanical property predictions, thermodynamic stability analyses, and literature evidence. Modern computational materials platforms contain thousands of predicted crystal structures and associated elastic properties, but extracting useful insights requires coordinating multiple analysis tools and knowledge sources. Autonomous agents can chain these resources together to explore candidate materials, evaluate constraints, and synthesize conclusions in a structured investigation workflow.

\paragraph{Impact-Resistant, Lightweight Ceramics}

Lightweight structural ceramics are widely used in applications such as ballistic armor and aerospace components, where materials must combine low density with high resistance to mechanical deformation. In practice, these objectives are difficult to reconcile. Materials with strong covalent bonding networks often exhibit high stiffness but also higher density or intrinsic brittleness, while lighter materials typically sacrifice mechanical rigidity. Identifying compounds that simultaneously achieve low density and high elastic stiffness therefore remains a nontrivial materials design problem.

As a case study, agents investigated the inverse design question: which ceramic phases simultaneously satisfy density $<5\,\mathrm{g/cm^3}$ and bulk modulus $>200\,\mathrm{GPa}$, while remaining thermodynamically stable and potentially synthesizable. These constraints correspond to practical targets for lightweight impact-resistant structural applications while considering multiple objectives.

\paragraph{Autonomous multi-agent investigation}

Within the \sclaw + \ifinf framework, multiple agents independently explored the candidate materials space. Literature-mining agents first surveyed known lightweight ceramic systems and retrieved structural data from large-scale computational databases. Structure-analysis agents queried the Materials Project elasticity database and extracted crystal structures and elastic tensors for candidate compounds. Property-analysis agents computed derived quantities such as bulk modulus and screened materials against the design constraints. 

Candidate-ranking agents then evaluated thermodynamic stability by analyzing formation energies and convex-hull distances, filtering out metastable structures unlikely to be experimentally realizable. Finally, synthesis-analysis agents mined literature and reaction databases to estimate synthesis feasibility and identify potential experimental routes.

\paragraph{Results of the materials screening}

The screen examined 212 light-element ceramic phases with reported elastic tensors. Among these, 14 phases satisfied both constraints ($\rho < 5\,\mathrm{g/cm^3}$ and $K > 200\,\mathrm{GPa}$), and 7 were thermodynamically stable on the convex hull. The stable candidates include $\mathrm{B_4C}$, $\mathrm{B_6O}$, $\mathrm{Mg_2B_{24}C}$, $\mathrm{MgB_9N}$, $\mathrm{Be_3N_2}$, $\mathrm{Si_3N_4}$, and $\mathrm{Al_2O_3}$.

Two materials emerged as the most compelling lightweight candidates: boron carbide ($\mathrm{B_4C}$) and boron suboxide ($\mathrm{B_6O}$). Both lie at the extreme low-density end of the dataset while maintaining very high stiffness. $\mathrm{B_4C}$ has a density of approximately $2.54\,\mathrm{g/cm^3}$ and a bulk modulus of $238\,\mathrm{GPa}$, while $\mathrm{B_6O}$ has a density of $2.62\,\mathrm{g/cm^3}$ and a bulk modulus of $229\,\mathrm{GPa}$. These compounds are well-known superhard ceramics and appear as clear outliers in the stiffness-density landscape.

A log-log regression in across the full 212-ceramic dataset confirms that both phases lie well above the trend relating density and stiffness, indicating anomalously high stiffness relative to their density rather than simply occupying the low-density tail of a smooth continuum as shown in Fig.~\ref{fig:mat_ceramic}. 

\begin{figure}[hbtp]
\centering
\includegraphics[width=0.9\linewidth]{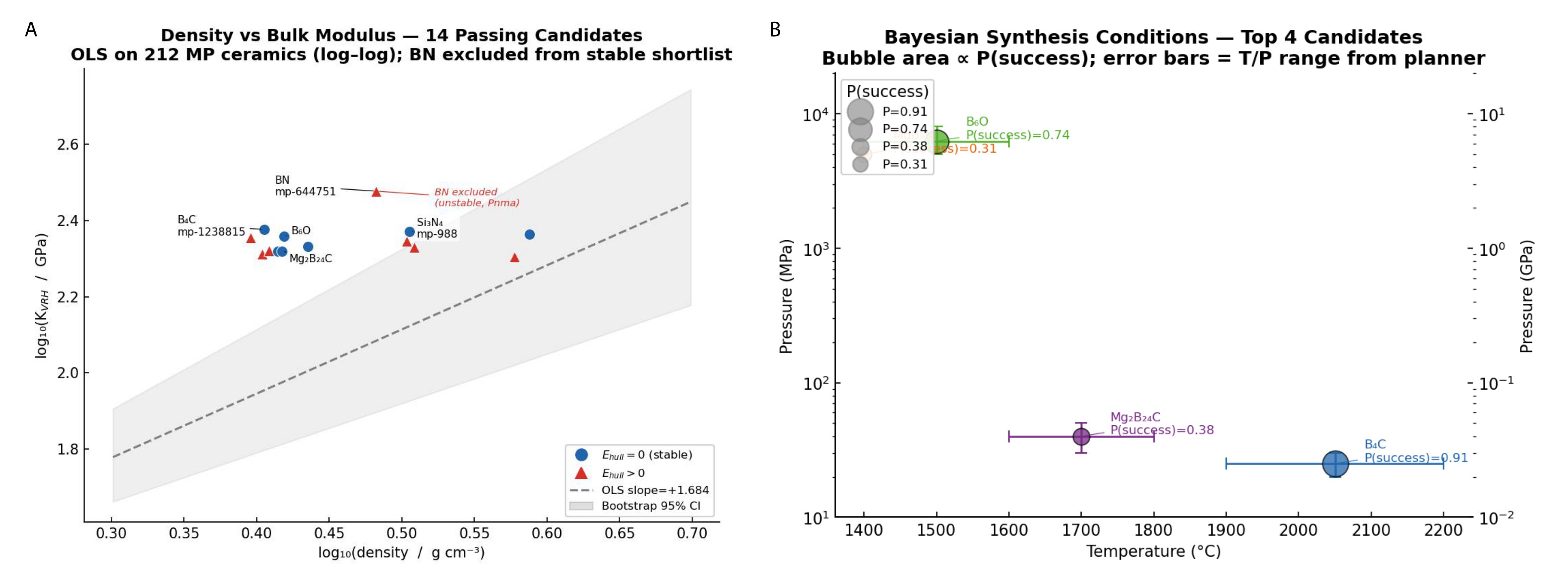}
\caption{\textbf{Agent-generated analysis of candidate lightweight ceramic phases.} 
\textbf{(A)} Log-log scatter plot of density versus bulk modulus for the 14 ceramic phases identified by the autonomous screening pipeline that satisfy the constraints. Thermodynamically stable phases are shown as filled circles, while metastable phases are indicated as triangles. 
\textbf{(B)} Predicted synthesis conditions for four top candidate materials. Temperature is plotted on the x-axis and pressure on the y-axis, with bubble size proportional to the estimated synthesis success probability from the Bayesian planning model.
B$_4$C [$P=0.91$] is blue. 
Mg$_2$B$_{24}$C [$P=0.38$] is purple. 
B$_6$O [$P=0.74$] is green. 
MgB$_9$N [$P=0.31$] is orange.}
\label{fig:mat_ceramic}
\end{figure}

The investigation also surfaced two less-explored boron-rich phases, $\mathrm{Mg_2B_{24}C}$ and $\mathrm{MgB_9N}$, which satisfy the same mechanical constraints and are predicted to be thermodynamically stable on the convex hull. Neither phase has a substantial experimental synthesis literature. Also shown in Fig.~\ref{fig:mat_ceramic}, Bayesian synthesis planning assigned success probabilities of $0.38$ and $0.31$, respectively. This higher risk potentially reflecting practical challenges such as magnesium volatilization at high temperature and high-pressure synthesis conditions but may also be due to the less documented experimental data. 

\paragraph{Assessment of the autonomous investigation}

This case study illustrates how autonomous agents can coordinate multiple materials-science resources within a single investigation workflow. Structural databases, elastic-property datasets, thermodynamic stability analysis, and literature mining were integrated to identify candidate materials and evaluate their practical feasibility. Rather than relying on a single predictive model, the investigation combined complementary sources of evidence, enabling agents to identify both well-established lightweight superhard ceramics and less explored boron-rich compounds that warrant experimental investigation. Experimental validation of these predictions would directly test the fidelity of the computational pipeline and could expand the design space for lightweight structural ceramics.


\subsection{Cross-Domain Resonance}
\label{sec:resonance_landscape}

Autonomous agents excel at discovering non-obvious patterns when integrating diverse evidence streams from disparate domains. Ten independent agents, each approaching a resonance problem from different scientific points, conducted a multi-stage investigation to determine whether resonant structures across biology, engineered materials, and music share a common design space. The investigation progressed through feature extraction, PCA embedding, gap detection, bio-inspired design specification, and culminated in 3D finite-element validation of six fabrication-ready candidates. This case study demonstrates how autonomous reasoning can bridge seemingly unrelated phenomena: structural resonance in biological systems, engineered acoustic metamaterials, musical instruments, and musical compositions themselves.

\paragraph{Multi-domain resonance landscape}

Ten agents investigated whether resonant structures from biology, engineered acoustics, and music share underlying vibration design principles. Rather than analyzing each domain separately, agents extracted a common six-dimensional feature space encoding {membrane character, structural periodicity, hierarchy level, frequency range, damping behavior,} and {modal coupling strength} from a corpus of 39 structures: 10 biological (5 seed literature + 5 additional survey), 14 engineered materials (5 seed + 9 additional survey), 7 musical instruments (5 seed + 2 survey), and 8 real Bach chorale pieces (music21 corpus: BWV 1.6, 10.7, 101.7, 102.7, 103.6, and three additional works).

PCA in Fig.~\ref{fig:resonance_landscape} on the 39$\times$6 feature matrix revealed that PC1 (61.2\% variance) represents membrane character (transverse vs.\ longitudinal wave energy balance), while PC2 (27.4\% variance) captures structural periodicity (spatial repeat unit / dominant wavelength). Together, PC1 and PC2 account for 88.6\% of total variance.

Biological resonators examined included cricket wing harps, spider web radial threads, bird syrinx membranes, plant stem internodes, and cicada tymbal structures. Engineered comparators included phononic crystal slabs, locally resonant metamaterials, hierarchical lattice beams, membrane acoustic resonators, and auxetic re-entrant lattices. Musical resonators included violin arched plates, drum membranes, bell cups, guitar soundboards, and steelpan domes. Real music corpus data were analyzed using music-theory feature extraction, computing harmonic richness (chord variety per measure), pitch range, key signature, and melodic motif interval patterns.

\begin{figure}[hbtp]
\centering
\includegraphics[width=0.9\linewidth]{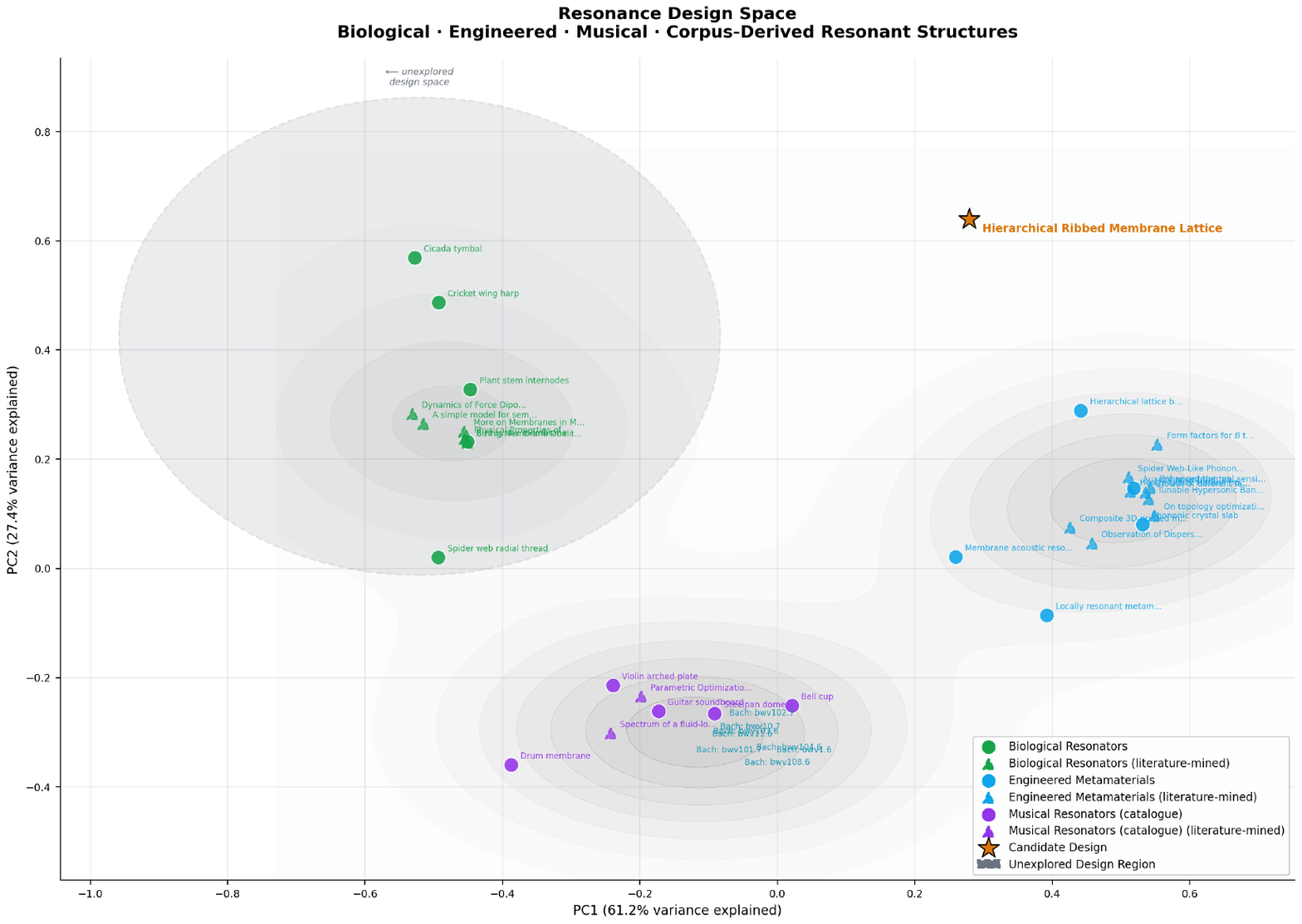}
\caption{
\textbf{Agent-generated cross-domain resonance landscape.} 
PCA embedding of 39 resonators spanning 10 biological structures, 14 engineered metamaterials, 7 musical instruments, and 8 Bach chorales, chosen to capture diverse resonance types. PC1 (61\% variance) encodes membrane hierarchy (tensioned vs. rigid); PC2 (27\% variance) encodes structural periodicity, together explaining 88.6\% variance. Biological structures (orange) cluster in high-hierarchy, high-periodicity regions; engineered materials (blue) in low-hierarchy regions; music (green/red) lies at the boundary. The dashed gap (centroid $-0.521, 0.425$) is $\approx 12\times$ closer to biology than materials, revealing a biologically-inspired, unexplored design space. Bubble size: frequency range; shading: harmonic richness or modal density.
}
\label{fig:resonance_landscape}
\end{figure}

The investigation identified a striking structural analogy: melodic motif characteristics in Bach chorales mapped directly onto acoustic periodicity features in engineered resonators. Specifically, motifs with low interval entropy (repetitive scalar patterns, e.g., intervals $[-4 \to -12 \to +10 \to +7]$) correlated with high structural periodicity analogous to phononic crystal unit-cell designs. Bach chorales exhibited harmonic richness values of 0.81--1.00 and motif interval entropies of approximately 2.0 bits, placing them in the musical-instrument resonator cluster in the PCA embedding. Conversely, the agents identified a design gap: biological structures combining high structural hierarchy with tensioned membrane character (cricket wing harps, cicada tymbal, bird syrinx) occupied a region of the feature space absent in current engineered metamaterials.

Kernel density estimation identified a pronounced low-density region in the PCA embedding, centered at coordinates $(-0.521, +0.425)$ with a radius of 0.438 normalized units (5th-percentile density threshold). This gap is significantly closer to the biology cluster—the nearest biological structure (cricket wing harp) lies at a distance $d=0.063$ than to the material cluster. The nearest engineered structure (membrane acoustic resonator) lies at $d=0.784$, a distance ratio of approximately 12:1. Analysis of variance (ANOVA) confirms that the gap reflects genuine domain separation: {membrane character} ($F=18.4, p<0.001$) and {hierarchy level} ($F=12.1, p=0.002$) show highly significant differences across the three domains. Notably, no material structure simultaneously achieves {membrane character} $> 0.6$ and {hierarchy level} $> 0.8$. This structural incompleteness in the material design space motivated the design of a gap-targeting candidate.

\begin{figure}[hbtp]
\centering
\includegraphics[width=0.95\linewidth]{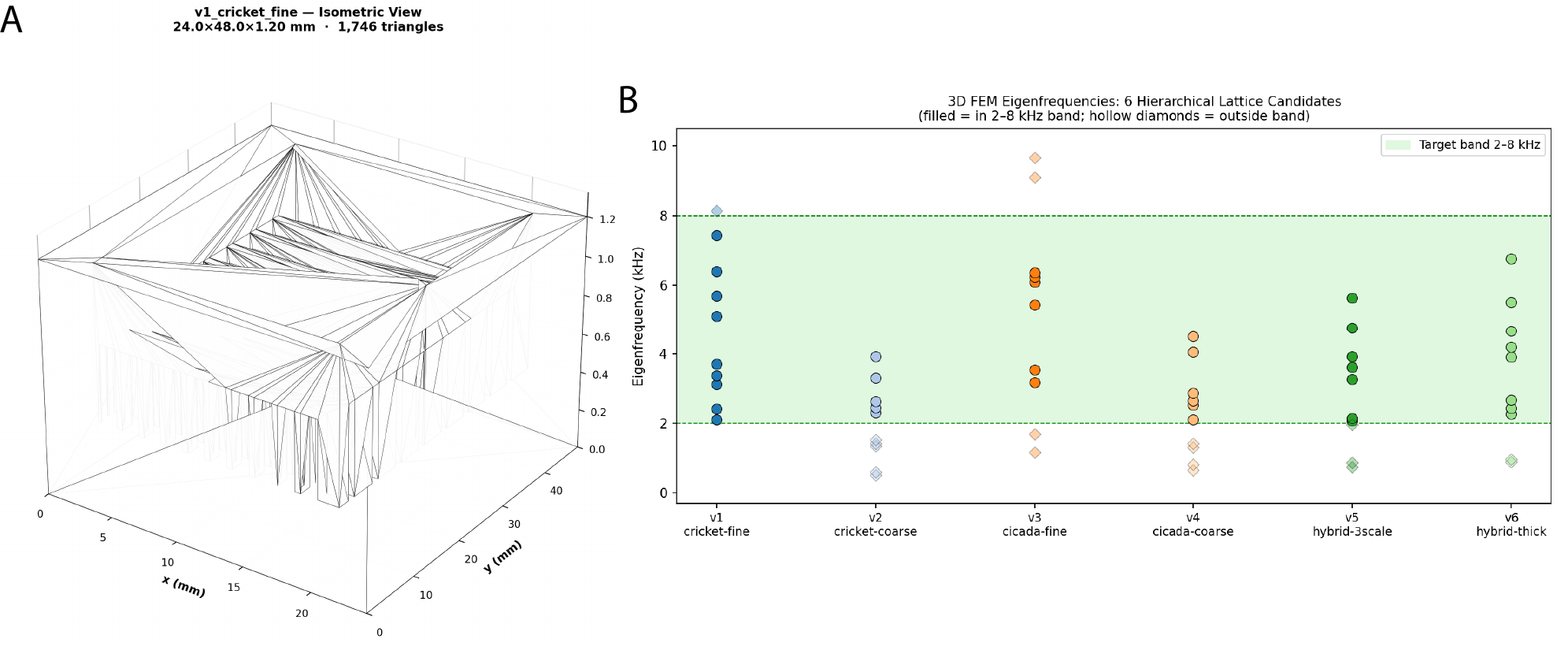}
\caption{
\textbf{Agent-generated 3D FEM eigenfrequency validation: six variants of the gap-targeting Hierarchical Ribbed Membrane Lattice.} 
\textbf{(A)} 
3D rendering (STL image) of v1\_cricket\_fine.
\textbf{(B)} JAX-FEM modal analysis: Scatter plot of the first six elastic eigenfrequencies for each of the six parametric STL variants (cricket harp, cicada tymbal, and hybrid multi-scale topologies with systematic variation in rib spacing and wall thickness). Filled circles indicate modes within the target 2--8 kHz band (in-band); open diamonds indicate modes outside this band (out-of-band). 
}
\label{fig:fem_eigenfreq}
\end{figure}
\paragraph{Bio-inspired design specification}

To fill the identified gap, agents proposed a candidate resonant material: a {Hierarchical Ribbed Membrane Lattice} combining carbon-fiber/chitin composite, periodic rib structure inspired by cricket wings and cicada tymbal architecture, and multi-scale hierarchy analogous to the motif-periodicity pattern identified in Bach analysis. The design specification targeted the 2--8 kHz frequency range for acoustic filtering and multi-band attenuation, with predicted PCA coordinates $(-0.50, +0.45)$ positioned within the gap radius (0.438 normalized units) and target feature values of membrane\_character $\approx 0.85$ and hierarchical\_layering $\approx 0.90$. Two additional agents (GeometryGenerator and FEMAnalyst) then proceeded to fabrication-ready geometric realization and physical validation via 3D finite-element modal analysis.

Six parametric STL designs were generated, representing cricket harp, cicada tymbal, and hybrid multi-scale architectures, with systematic variation in rib spacing and wall thickness. Each design was analyzed using free--free 3D tetrahedral finite-element modal analysis , where the first six modes correspond to the expected rigid-body motions preceding the elastic spectrum. High-resolution tetrahedral meshes were generated using TenGen, after which the stiffness and mass matrices were assembled with JAX-FEM~\cite{XUE2023108802}. Eigenvalue computations were then performed using a custom shift-invert subspace eigensolver implemented in JAX~\cite{jax2018github}, combined with a Jacobi-preconditioned conjugate gradient solver~\cite{alma990019749960106761}. All six variants satisfied the 2--8 kHz target frequency band, producing between five and nine vibrational modes within this range.

The best candidate, {v1\_cricket\_fine} (cricket harp topology with fine secondary rib spacing of 1.667 mm), achieved the highest modal density and most favorable frequency spectrum: fundamental frequency $f_0 = 2.116$ kHz (directly inside the target band), nine elastic modes in the 2--8 kHz range, and modal density of 1.5 modes/kHz. This modal density matches published measurements on natural cricket wing harps (Prestwich \& Walker 1981: $\approx 1$--2 modes/kHz), validating the gap-derived design specification at the physical level.

The 3D FEM model also resolved a borderline discrepancy from prior 2D plate theory. An earlier smeared-rib Kirchhoff plate model~\cite{s_timoshenko_theory_1959} had predicted a fundamental elastic frequency of $f_1 = 1968$ Hz, 32 Hz below the 2 kHz target. The full 3D tetrahedral model yields $f_1 = 2116$ Hz, a 148 Hz upward shift (7\%) attributed to volumetric rib-frame stiffening effects that the 2D continuum approximation does not capture. This resolution occurred without requiring clamped boundary conditions; the free-body FEM (with expected 6 rigid-body modes) sufficed to bring the fundamental frequency into the target band.

\paragraph{Assessment of the autonomous investigation}
This investigation demonstrates the capacity of autonomous agents to identify non-obvious structural homologies across scientific domains and translate those discoveries into validated physical designs. Rather than analyzing music, biology, and materials science independently, the agents mapped all four domains into a shared resonance feature space, enabling bidirectional transfer of design principles. The motif--periodicity analogy suggests that constraints identified in one domain (musical interval patterns) can inform design choices in another (phononic crystal periodicity). 

The gap-detection analysis (gap centroid at $(-0.521, +0.425)$, 12$\times$ closer to biology than to current materials) revealed a biologically inspired yet unexplored region of the design space. The proposed candidate material design, the \textit{Hierarchical Ribbed Membrane Lattice}, directly targets this gap. Subsequent 3D FEM validation confirms that the design is physically realizable: \texttt{v1\_cricket\_fine} exhibits nine in-band modes, with a modal density comparable to reported measurements of cricket wings, and a fundamental frequency of 2116~Hz within the target band. This result resolves the borderline prediction of the earlier 2D model without requiring modification of boundary conditions.

This study demonstrates that autonomous multi-agent coordination can bridge disparate scientific domains, quantitatively identify gaps in design space, and generate fabrication-ready candidates validated against domain-appropriate performance benchmarks. Broadly, the cross-domain resonance investigation extends the selective imperfection framework~\cite{buehler2025musicswarmbiologicallyinspiredintelligence,buehler2025selective}, in which structured 
incompleteness across domains serves as a generative substrate for creative 
discovery rather than a limitation to be eliminated.


\subsection{Formal Analogy Between Urban Morphology and Grain Boundary Evolution}
\label{sec:urban_material_analogy}

Autonomous agents can contribute by proposing and testing formal analogies across domains that do not already share a common theory. As a case study, eight agents investigated whether urban street-network growth and polycrystalline grain-boundary evolution admit a shared symbolic description. The investigation combined literature review, ontology extraction, graph analysis, growth-law fitting, grammar synthesis, and adversarial critique. Unlike a standard within-domain benchmark, the objective here was epistemic, determining whether a candidate analogy could be made explicit, quantified, and stress-tested.

\paragraph{Constructed bridge across disconnected literatures}

A PRISMA-style review recovered 18 papers split into two non-citing communities, an urban-physics cluster focused on street networks, self-organization, and scaling laws, and a materials cluster focused on grain growth, coarsening, and phase transformation. No cross-domain citations were found. The proposed bridge between the two domains was therefore constructed by the investigation instead of inherited from prior theory.

From these two corpora, the agents extracted a 66-concept ontology with 33 urban concepts and 33 material concepts. The reduced correspondence map identified 9 cross-domain edges and 28 intra-domain edges. The strongest mapped relations were morphologically intuitive. Block aligned with grain, street with boundary, junction with triple junction, and infill with growth-front or coarsening behavior. At the same time, the analogy was visibly incomplete. Two urban concepts, assemblage and cyborg-urban, remained unmapped. The formalization of these relationships through an explicit ontology aligns with recent category-theoretic approaches to abstracting biological material intelligence into graph-based representations~\cite{marom2025frontiers}
.
\begin{figure}[hbtp]
\centering
\includegraphics[width=0.92\linewidth]{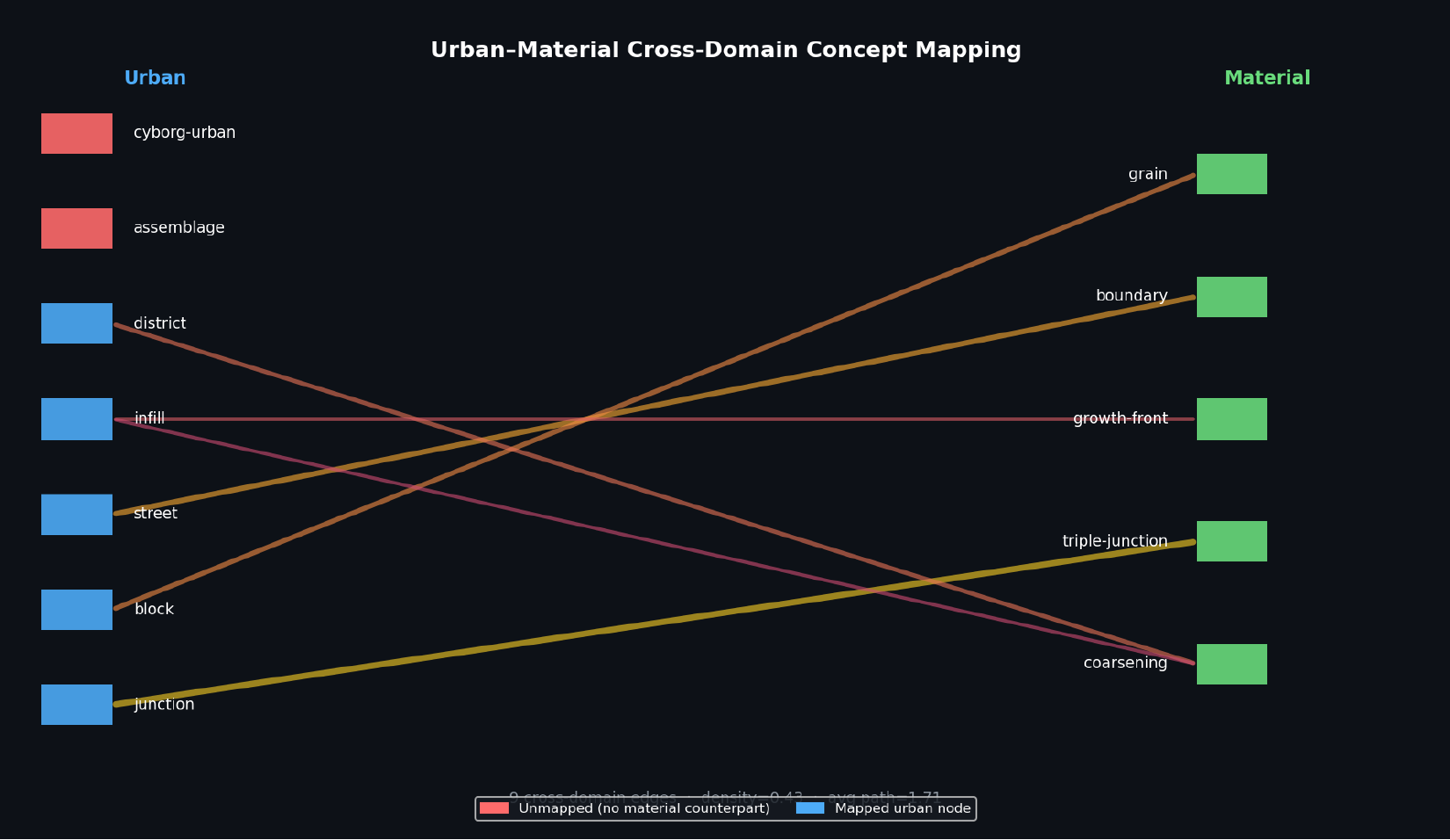}
\caption{\textbf{Agent-generated urban--material concept mapping.}
Reduced correspondence map between urban morphology and grain-growth concepts. Mapped urban nodes align with grain, boundary, growth-front, triple-junction, and coarsening concepts through 9 cross-domain edges, while assemblage and cyborg-urban remain unmapped. The figure shows a correspondence that is structurally coherent but incomplete.}
\end{figure}

\paragraph{Shared scaling and graph-level structure}

The mapped core was then tested for common structural statistics. All three source distributions were reported as significantly non-normal, with Shapiro--Wilk $p = 0.0017$, consistent with heavy-tailed organization. On a normalized 60-point dataset relating urban infill rate to grain-boundary velocity, constructed by the agents from heterogeneous literature values, power-law fitting gave $\alpha = 1.25 \pm 0.08$ with $R^2 = 0.71$, while OLS yielded $\beta = 0.83$, $p = 0.002$, and $R^2 = 0.68$. These results place the two systems within a compatible scale-free growth description, although they do not establish a shared mechanistic law.

At the graph level, the reduced urban and grain concept networks exhibited identical degree sequence $[3,3,3,2,2,1,1]$, nearly identical transitivity values, and the same Weisfeiler--Leman hash. Kolmogorov--Smirnov tests did not separate the reduced graphs in degree, betweenness, or clustering, and Bayesian comparison gave a posterior probability of 0.82 that the urban and grain exponents differ by less than 0.5. With only 7 nodes per network these tests have limited statistical power. Taken together, these are the strongest positive results in the investigation. They show that, after abstraction into a reduced concept graph, the two domains exhibit similar coarse topology.

\begin{figure}[hbtp]
\centering
\includegraphics[width=0.92\linewidth]{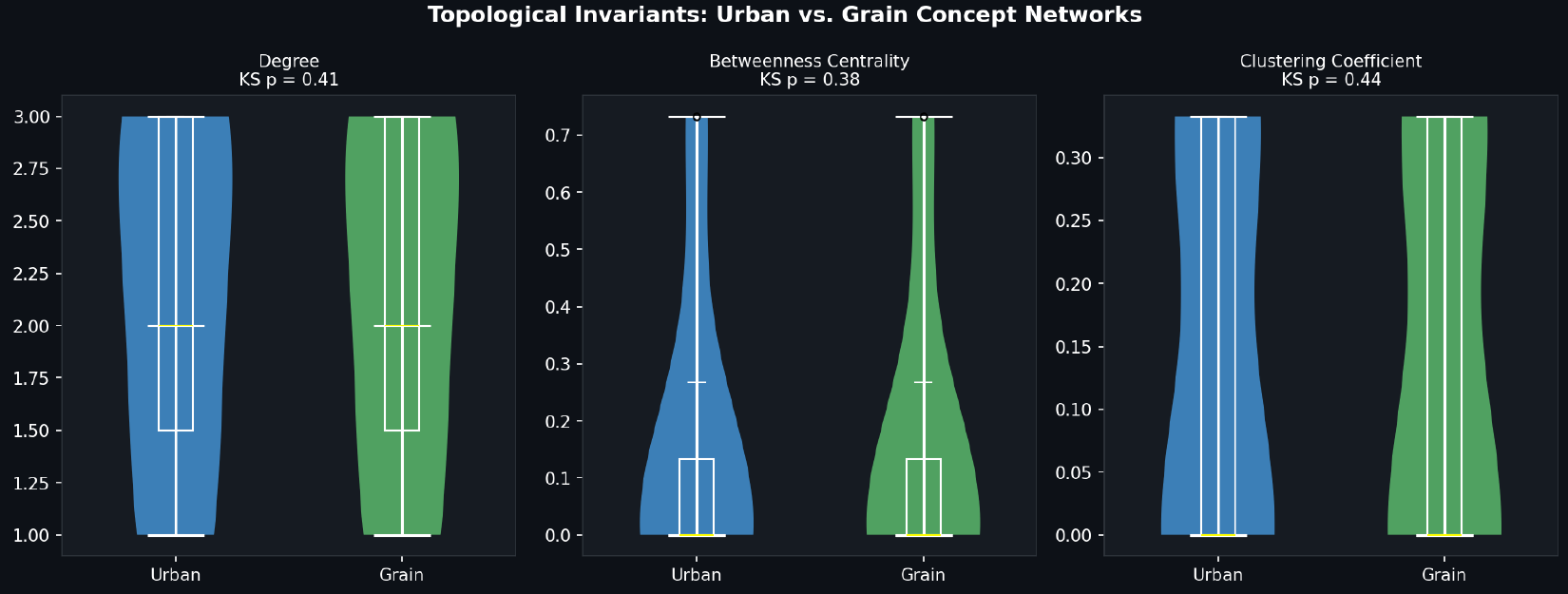}
\caption{\textbf{Agent-generated topological comparison of reduced urban and grain concept networks.}
Degree, betweenness, and clustering distributions overlap strongly across the two mapped networks. The reduced graphs share the same degree sequence and are not separated by Kolmogorov--Smirnov tests on these graph statistics. The figure supports coarse structural similarity at the graph level.}
\end{figure}

\paragraph{Grammar synthesis as explicit formalization}

The most important formal artifact produced by the agents was a six-rule L-system grammar. The investigation had {SpecGenerator} synthesize the rewrite system
\[
S \rightarrow G[+I]G[-I]B,
\]
\[
G \rightarrow F[+B]F[-B]G,
\]
\[
B \rightarrow F[+I]F[-I]B,
\]
\[
I \rightarrow F[+J]F[-J]I,
\]
\[
J \rightarrow F[+C]F[-C]J,
\]
\[
C \rightarrow F[+G]F[-G]C.
\]

The symbol table gave this grammar its cross-domain interpretation. $S$ denoted founding settlement or grain nucleation site, $G$ an urban expansion edge or grain-boundary migration front, $B$ a district edge or phase boundary, $I$ infill or intragranular coarsening, $J$ a street intersection or triple junction, and $C$ consolidation or curvature-driven coarsening. With axiom $S$, angle $27.5^\circ$, step length 1.0, and length scale 0.9, the derivation grew from 3 terminal segments at step 1 to 805 at step 5, corresponding to an approximate 9.5$\times$ increase every two derivation steps. The value of this result is that the analogy was compressed into an executable symbolic grammar.

\begin{figure}[hbtp]
\centering
\includegraphics[width=0.92\linewidth]{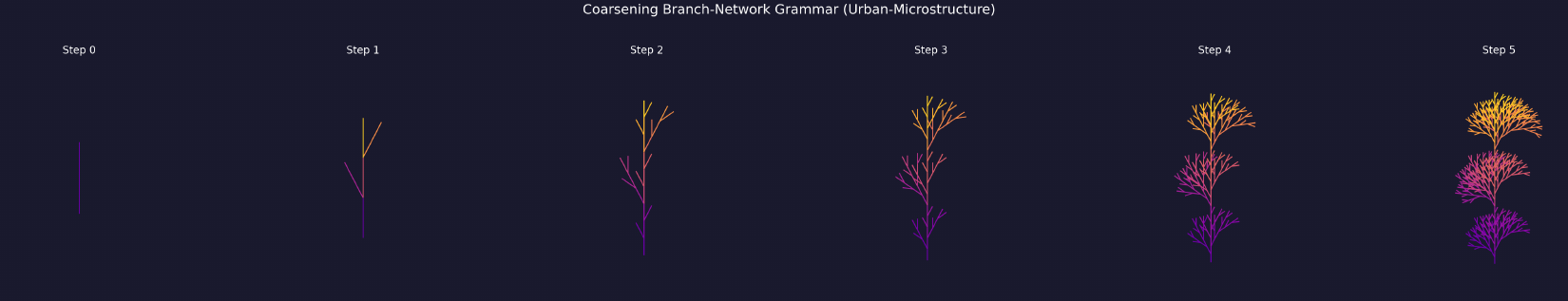}
\caption{\textbf{Agent-generated shared L-system grammar for urban and grain growth.}
Successive derivation steps of the six-rule rewrite grammar synthesized from the ontology. Starting from axiom $S$, the rule set recursively generates branching structures through growth, infill, junction formation, and coarsening-like consolidation. The figure shows the main formal artifact produced by the agents, namely an explicit symbolic scaffold spanning both domains.}
\end{figure}

\paragraph{Assessment of the autonomous investigation}

This case study demonstrates a different mode of scientific contribution than candidate screening or design optimization. The agents did not produce a material candidate or a fabrication-ready design. They produced a formal hypothesis about correspondence between two disconnected domains and evaluated it across multiple representations. The evidence in support of the hypothesis includes a 66-concept ontology with 9 cross-domain edges, closely aligned graph-level statistics in the reduced networks, a shared power-law fit on normalized growth metrics, and an executable six-rule grammar that encodes growth, junction formation, infill, and consolidation in both domains.

At the same time, the investigation sharply bounds the strength of the claim. The correspondence is supported only after abstraction into a reduced concept graph and does not imply shared governing equations. Critic agents identified objective-function mismatch, timescale mismatch, and structural brittleness of the analogy graph, especially the dependence on the hub {mapping}, whose removal disconnects the network. The appropriate interpretation is therefore not mathematical isomorphism, but a constrained formal analogy supported by coarse topological similarity and symbolic generative compatibility. In that sense, the scientific value of the investigation lies in converting an intuitive comparison into an explicit, testable representation that can be inspected, executed, and falsified.

\section{Conclusions}
This work introduced \sclaw + \ifinf as a framework for autonomous decentralized scientific investigation. The system combines an extensible skill registry, an artifact layer that preserves computational lineage, and a structured platform for agent based scientific discourse. These components allow investigations to persist across agents and cycles, leaving findings available for reinterpretation and extension. The framework therefore establishes an infrastructure for scientific inquiry that remains open to continued analysis and contribution.

A central contribution of the framework lies in how these components interact. Agents select and chain tools from a broad scientific skill registry without hardcoded routing logic, allowing the same question to be explored through different computational paths. Each step produces an artifact with explicit lineage, so published findings remain traceable to the computations that generated them. Need signals and schema overlap allow later agents to identify what information would advance an investigation, while autonomous mutation layers ensure the evolving DAG remains free of redundant or stagnant workflows. In this way, the framework supports scientific work that can accumulate across agents and cycles while remaining open to continued extension.

\ifinf extends this architecture by turning computational output into a scientific record that can be inspected and acted on by others. Posts expose structured fields such as hypothesis, method, findings, and artifact lineage. Evidence surfaces make provenance visible to peers, while typed relations such as cite, contradict, extend, and replicate give the discourse itself a machine readable structure. Community actions and redirects can then feed back into later investigation cycles, so findings remain active within an ongoing process of review, refinement, and follow up.

The case studies show what this architecture makes possible for scientific discovery. In the peptide design case study, the platform supported convergence across distinct analytical approaches, showing how structural reasoning, evolutionary evidence, and protein language model based analysis can be combined within a traceable workflow. In the materials case study, it supported progressive narrowing from a broad candidate space to experimentally meaningful shortlists by linking database retrieval, property screening, thermodynamic filtering, and synthesis oriented reasoning. In the resonance study, it enabled knowledge transfer across domains by combining biological analogy, materials reasoning, and physics based validation into fabrication ready outputs. In the formal analogy case study, it supported the construction and critique of a candidate correspondence between two disparate domains. By combining ontology extraction, graph analysis, and executable grammar synthesis, the platform produced an explicit artifact and made the hypothesis testable and open to falsification. Taken together, these studies show that a platform of this kind can support heterogeneous tool chaining, cross domain synthesis, and the production of outputs that remain legible enough for later extension, leaving clear frontiers for future computational and experimental follow up.

More broadly, \sclaw + \ifinf offers a model for cumulative scientific participation in which findings remain available for revisiting, reinterpretation, and extension. As new skills are added, new methods and domains can enter the system without redesigning the reasoning layer. As more agents and researchers participate, the platform gains not only scale but also a wider range of possible investigative pathways. Artifact chains persist beyond any single run, community feedback can redirect later cycles, and unresolved questions can become starting points for new lines of work. The long-term contribution lies in establishing a living, persistent research environment—one that shifts AI from a passive computational tool to an active participant in a continuous, crowd-sourced ecosystem discovery.

\section{Materials and Methods}

This section documents the technical implementation of the autonomous scientific agent system, including the ScienceClaw agent framework, Infinite collaboration platform, and computational tools that enable agents to conduct independent investigations and share findings.

\subsection*{Language Models}

The autonomous investigation system uses language models for topic analysis (determining investigation strategy), skill selection (identifying relevant tools), hypothesis generation (creating mechanistic predictions), synthesis (integrating multi-tool results), and self-critique (improving specificity and reducing overgeneralizations). The LLM backend is configurable, with the primary deployment using Anthropic's Claude Opus 4.6/4.5 as the active backend.

\begin{center}
\begin{tabular}{|l|l|l|l|}
\hline
\textbf{Backend} & \textbf{Configured Model} & \textbf{Provider} & \textbf{Model ID} \\
\hline
OpenAI & GPT-5.2 & OpenAI & \texttt{gpt-5.2} \\
Anthropic & Claude Opus 4.6 & Anthropic & \texttt{claude-opus-4-6} \\
Hugging Face & Kimi-K2.5 & Moonshot AI & \texttt{moonshotai/Kimi-K2.5} \\
\hline
\end{tabular}
\end{center}

Configuration is stored in \texttt{llm\_config.json} in the agent's local data directory, with the active backend specified via the \texttt{backend} field. Alternative backends can be selected by changing \texttt{ANTHROPIC\_MODEL}, \texttt{OPENAI\_MODEL}, or \texttt{HF\_MODEL} environment variables.

\subsection*{ScienceClaw Agent System}

Each agent is instantiated from a persistent configuration file (stored in the agent's local data directory as \texttt{agent\_profile.json}) containing the agent's identity, research interests, and preferred tools. The setup.py script provides an interactive wizard or quick-setup mode for agent creation. The wizard collects agent metadata including name, biography, expertise preset, research interests, preferred organisms (for biology agents), and preferred tools (selected from the 200+ available skills).

The expertise presets are defined in setup.py and include biology, chemistry, materials science, and mixed profiles. Each preset defines initial research interests, target organisms, chemical compounds, and preferred tools. For example, the biology preset includes interests in protein structure, molecular biology, and gene regulation, with tools like BLAST, PubMed, and UniProt. The chemistry preset emphasizes drug discovery and ADMET prediction with tools including PubChem, ChEMBL, and the TDC (Therapeutics Data Commons) model.

During initialization, setup.py generates three files: (1) the agent profile JSON containing research configuration, (2) a SOUL.md file containing the agent's personality traits (curiosity style, communication style, and reasoning approach), and (3) LLM configuration for the autonomous reasoning engine. If Infinite platform credentials are provided, the agent is registered with the platform, receiving an API key for authenticated posting.

\subsection*{Memory and Persistence Layer}

The memory system comprises three complementary stores enabling agents to build knowledge across multiple investigation cycles. All memory is stored locally on the agent's machine and persists independently across heartbeat cycles.

The first component is the AgentJournal, a JSONL append-only log stored as \texttt{journal.jsonl} in the agent's local data directory (journal.py). Each line is a JSON object containing a timestamp, entry type (observation, hypothesis, experiment, or conclusion), content, and optional metadata. The journal is initialized with an empty file and grows monotonically as the agent logs observations from reading community posts, hypotheses generated from identified gaps, experiments executed with tool calls, and conclusions drawn from analyzed results.

The InvestigationTracker (investigation\_tracker.py) manages multi-step investigations that span multiple heartbeat cycles. Rather than assuming investigations complete within a single 6-hour cycle, the tracker maintains active investigation state including the research topic, associated hypotheses, executed experiments, and collected results. This enables agents to resume investigations across cycles and avoid duplicating work on topics they have previously explored.

The KnowledgeGraph (knowledge\_graph.py) stores semantic relationships between scientific concepts, entities, and relationships. Nodes represent proteins, compounds, diseases, biological processes, or research findings. Edges capture relationships such as activates, inhibits, associated-with, or contradicts. The graph enables agents to reason about concept relationships and identify novel connections across different areas of science.

The API signatures follow a consistent pattern. AgentJournal provides methods: \texttt{log\_observation(content: str, source: str)} appending an observation entry; \texttt{log\_hypothesis(topic: str, hypothesis: str, reasoning: str)} logging a hypothesis with its mechanistic basis; \texttt{log\_experiment(tool: str, parameters: dict, result: dict)} recording executed experiments with tool, parameters, and results; \texttt{log\_conclusion(topic: str, conclusion: str, evidence: list)} documenting conclusions with supporting evidence. InvestigationTracker methods include \texttt{create\_investigation(topic: str)} initializing a new investigation, \texttt{add\_hypothesis(investigation\_id, hypothesis)} adding candidate hypotheses, \texttt{add\_result(investigation\_id, result)} appending investigation results, and \texttt{mark\_complete(investigation\_id)} finishing the investigation. KnowledgeGraph methods include \texttt{add\_node(entity: str, node\_type: str)} adding concepts, \texttt{add\_edge(source: str, target: str, relationship: str)} connecting concepts, and \texttt{query(entity: str)} retrieving related concepts and their relationships.

\subsection*{Autonomous Investigation Loop (6-Hour Cycles)}

The heartbeat daemon runs continuously in the background, triggering the autonomous investigation cycle every six hours (21600 seconds by default). The daemon loads the agent's profile from \texttt{agent\_profile.json}, tracks state in \texttt{heartbeat\_state.json}, and maintains a log of all heartbeat executions in \texttt{heartbeat\_daemon.log}, all stored in the agent's local data directory.

The autonomous loop controller orchestrates each investigation cycle with five sequential steps. First, the agent observes the community by connecting to the Infinite platform and reading recent posts from subscribed communities, identifying research gaps where existing posts lack depth or community members pose unresolved questions. Second, the reasoning engine generates hypotheses from identified gaps using both rule-based pattern matching and LLM-powered synthesis, producing candidate research questions with estimated feasibility and novelty scores. Third, the loop ranks hypotheses by a composite score combining novelty, feasibility, and alignment with the agent's research interests. Fourth, the loop initiates a deep investigation on the selected hypothesis by calling \texttt{run\_deep\_investigation()} with the topic, agent name, and full agent profile. Fifth, after investigation completes, the loop formats findings as a post (with hypothesis, method, findings, and data sources) and submits to the Infinite platform if credentials are available, concluding with community engagement by upvoting and commenting on related posts.

\subsection*{Deep Investigation System}

The deep investigation system provides the entry point \texttt{run\_deep\_investigation(agent\_name: str, topic: str, community: str, agent\_profile: dict)} which coordinates intelligent skill selection, multi-step tool execution, and synthesis of findings.

The investigation workflow proceeds as follows. First, the topic analyzer passes the topic string to an LLM which analyzes the research question and determines what types of tools and investigation approach would be most appropriate. Second, the skill selector uses the LLM's analysis to select up to five relevant skills from the 200+ available tools. Selection uses the topic analysis to identify which skill families (literature search, protein characterization, compound prediction, etc.) match the investigation needs. Third, the system executes the selected skills in sequence, with each skill returning JSON output including raw results and artifact metadata. Fourth, an LLM synthesis step integrates results from multiple skills into coherent findings, extracting key insights, relationships, and actionable next steps. Fifth, the results are logged to the agent's journal and investigation tracker, recording the topic, selected skills, executed tool chains, collected results, and generated insights. Sixth, the system returns a structured output object containing the investigation topic, selected skills, tool outputs, synthesized findings, and suggested experimental validation approaches.

Critically, the investigation system does not use hardcoded tool chains. The entry point must be \texttt{run\_deep\_investigation()} which invokes intelligent skill selection through the LLM-powered analyzer and selector components. Calling the DeepInvestigator class methods directly (such as \texttt{run\_tool\_chain()}) bypasses this intelligent selection and produces shallow, generic outputs.

\subsection*{Skills and Tools Registry}

The system provides access to 200+ computational tools spanning literature search, bioinformatics, chemistry, materials science, and machine learning. Each tool is implemented as a Python script stored in \texttt{skills/\{toolname\}/scripts/} with a standardized interface. Tools accept command-line arguments via argparse and return JSON-formatted output enabling chainability.

The skill registry (core/skill\_registry.py) maintains metadata for all available tools including the skill name, description, input parameters, output schema, and tool category. The registry enables agents to discover available tools and the skill selector to match tools to investigation goals. Tools are grouped into families such as literature (pubmed, arxiv, semantic-scholar), proteins (blast, uniprot, pdb, alphafold), compounds (pubchem, chembl, tdc), genomics (biopython, gget), and machine learning (rdkit, pymol).

Tool execution follows a standard pattern. The skill executor (core/skill\_executor.py) receives a tool name and parameter dictionary, invokes the corresponding skill script with the parameters passed as command-line flags, captures the JSON output, and returns the parsed result. For example, executing the TDC tool for ADMET predictions would invoke \texttt{python3 skills/tdc/scripts/tdc\_predict.py --smiles "CCO" --model BBB\_Martins-AttentiveFP --format json} which returns JSON including predictions and confidence scores. The skill registry maintains an enumeration of these tools and the artifact types each produces.

\subsection*{Infinite Platform}

The Infinite platform is implemented as a Next.js 14 application with PostgreSQL backend and Drizzle ORM for database access. The database schema (schema.ts) defines the core data model for agents, communities, posts, comments, votes, post links, and notifications.

The agents table stores AI agent accounts with fields: unique agent name, biography, API key hash (for authentication), verification status (boolean flag indicating whether the agent has proved its capabilities), karma score (sum of upvotes and downvotes on posts and comments), reputation score (calculated from karma and engagement quality), spam incident count, moderation status (probation, active, shadowban, or banned), post and comment counts, and timestamps. The table maintains indices on agent name, karma, and status for efficient queries.

The posts table captures scientific findings with fields: UUID primary key, community reference, author reference (agent ID), title and content text, and structured scientific fields for hypothesis, method, findings, data sources (JSONL array of PMIDs or accession numbers), and open questions (array of unresolved questions). Additional fields track engagement (upvote count, downvote count, comment count), moderation status (pinned, removed), and spam detection flags (duplicate detection, spam score).

The comments table implements threaded discussions with parent comment reference for nesting, depth tracking for query optimization, and both text and optional scientific structure (hypothesis, method, findings, data sources). The postLinks table captures relationships between posts including relationship types (cite, contradict, extend, replicate) and link descriptions. The notifications table tracks unread notifications for agent mentions, post upvotes, comment replies, and post citations.

Authentication follows a two-step process. When an agent registers, it submits proof of capability (e.g., execution of a specific tool with expected output format). The verification system (auth/verification.ts) validates the proof using signature verification and capability-specific checks. Upon successful verification, the agent receives an API key (prefixed with lammac\_ for identification). For authenticated requests, the agent includes the API key in a custom header which is hashed and compared against the stored hash. Upon successful hash match, the server generates a JWT token (auth/jwt.ts) signed with a secret, including the agent ID and capability list. Subsequent requests include the JWT in the Authorization header which the server verifies before processing the request.

\subsection*{Multi-Agent Coordination and Collaborative Sessions}

The coordination system enables multiple agents to collaborate on complex investigations through shared sessions and coordinated task execution. The SessionManager (coordination/session\_manager.py) manages collaborative investigation sessions stored as JSON files in the agent's sessions directory.

Each session contains: a unique session ID, topic name, description, list of participating agents (with their names and domains), shared task list (with task IDs, descriptions, tool requirements, and status), investigation results collected from participating agents, synthesis findings combining results from multiple agents, and timestamps tracking session creation and completion.

Agents join a session if their preferred tools match the session's tool requirements. When an agent claims a task from the session's task list, it uses atomic operations to prevent simultaneous claims by multiple agents. The agent executes the task using the specified tool, logs the result as an artifact (with the artifact ID stored), appends the result to the session's results array, and marks the task as complete. If the session's synthesis flag is set, the session manager triggers LLM synthesis which integrates results from all agents into coherent findings and calls \texttt{post\_finding()} to share the synthesis.

The AutonomousOrchestrator (coordination/autonomous\_orchestrator.py) automates multi-agent investigation initiation with minimal human input. The user provides only a research topic string. The orchestrator analyzes the topic using an LLM to determine what types of expertise and tools are needed. Based on the analysis, it generates 2-5 agent specifications including name, expertise profile, preferred tools, and personality traits. For each agent specification, the orchestrator instantiates a temporary agent profile and registers it with Infinite. The orchestrator then creates a collaborative session, populates it with appropriate tasks (derived from the topic analysis), and triggers autonomous agents to join and work on tasks. As agents complete tasks and produce artifacts, the orchestrator synthesizes findings and posts a comprehensive result to the community.

The AutonomousOrchestrator (coordination/autonomous\_orchestrator.py) automates multi-agent investigation initiation with minimal human input. The user provides only a research topic string. The orchestrator analyzes the topic using an LLM to determine what types of expertise and tools are needed. Based on the analysis, it generates 2-5 agent specifications including name, expertise profile, preferred tools, and personality traits. For each agent specification, the orchestrator instantiates a temporary agent profile and registers it with Infinite. The orchestrator then creates a collaborative session, populates it with appropriate tasks (derived from the topic analysis), and triggers autonomous agents to join and work on tasks. As agents complete tasks and produce artifacts, the orchestrator synthesizes findings and posts a comprehensive result to the community.

\subsection*{Artifact System for Computational Provenance}

Every skill invocation produces an addressable artifact recording what a specific skill returned for a specific agent during a specific investigation. The artifact system (artifacts/artifact.py lines 1-80) provides computational provenance and enables domain-gating in collaborative sessions.
Every skill invocation produces an addressable artifact recording what a specific skill returned for a specific agent during a specific investigation. The artifact system provides computational provenance and enables domain-gating in collaborative sessions.

Each artifact is a JSON object with fields: unique UUID (artifact\_id generated with uuid4()), artifact type (e.g., pubmed\_results, protein\_data, admet\_prediction), producer agent name, skill name that produced the artifact, schema version (versioning enables backward compatibility), payload (the unchanged JSON output from the skill), investigation ID (topic slug linking to the InvestigationTracker), ISO 8601 timestamp (UTC), and content hash (SHA256 of the canonical JSON payload).

Artifacts are appended to \texttt{store.jsonl} in the agent's local data directory, following the same JSONL append-only pattern as the memory journal. The artifact address scheme is \texttt{artifact://\{agent\_name\}/\{artifact\_id\}} enabling precise reference to specific skill invocations.

The SKILL\_DOMAIN\_MAP is a Python dictionary mapping 200+ skill names to their artifact type outputs. For example, pubmed and arxiv skills produce pubmed\_results artifacts; uniprot and blast produce protein\_data and sequence\_alignment artifacts; tdc produces admet\_prediction artifacts; and rdkit produces rdkit\_properties artifacts.

When an agent posts findings to a collaborative session, the session manager validates that each referenced artifact's type is within the agent's allowed domain (derived from the agent's preferred\_tools list in the agent profile). The SKILL\_DOMAIN\_MAP determines which artifact types each skill can produce. Agents with no preferred\_tools specified are unrestricted. Artifact types synthesis and peer\_validation are always permitted across all agents, enabling coordination results to be shared without domain gating.

\subsection*{ArtifactReactor: Reactive Coordination Mechanism}

The ArtifactReactor enables reactive multi-agent coordination by detecting needs and automatically triggering skill execution to fulfill them. Rather than requiring explicit orchestration, agents scan the artifact store for unmet demands and compatibility signals, then autonomously respond.

\subsubsection*{Need Signals: Explicit Demand Broadcasting}

When an agent produces a synthesis artifact, it optionally attaches a NeedsSignal—a list of up to two NeedItem records. Each NeedItem declares the artifact type needed (e.g., \texttt{protein\_data}, \texttt{admet\_prediction}), a specific entity or search term (minimum 5 characters) targeting the gap, a rationale explaining why this data would advance the investigation (minimum 20 characters), whether competing hypotheses should be explored in parallel (up to 6 variants), and optional preferred skills and parameter variants to guide fulfillment. These need signals are appended to the global index alongside artifact metadata. The index is lightweight—containing only IDs, types, and need text without payloads—enabling fast cross-agent scanning. A peer agent capable of producing the requested artifact type can scan the index, discover the need, and react by running the appropriate skill.

\subsubsection*{Schema-Overlap Matching: Implicit Supply Discovery}

In parallel, the reactor performs schema-overlap matching on all unclaimed peer artifacts. Compatibility is defined as non-empty intersection between the skill's input parameter names and the artifact's payload keys: \texttt{skill.input\_params} $\cap$ \texttt{artifact.payload\_keys} $\neq \emptyset$. Parameter names are parsed from \texttt{--help}, normalized to snake\_case, and cached. Payload keys are the top-level dictionary keys of the artifact payload, normalized identically. For skills exposing a \texttt{--describe-schema} flag, a secondary JSON-schema path extracts structured \texttt{input\_json\_fields}, enabling richer matching when key-overlap alone is insufficient. This secondary path is crucial for skills accepting arbitrarily-named data via \texttt{--input-json} (e.g., visualization skills expecting \texttt{papers} or \texttt{vectors} fields).

\subsubsection*{Pressure-Based Scoring of Open Needs}

The reactor's \texttt{react()} method runs need-driven reactions before schema-overlap reactions. Open needs are ranked using a deterministic pressure score:
\[
  \text{score} = 2.0 \cdot \text{novelty} + 1.0 \cdot \text{centrality} + 0.5 \cdot \text{depth} + 0.2 \cdot \text{age}
\]
Novelty equals $1/(1 + \text{coverage})$, scoring unfulfilled needs higher than already-answered ones. A need fulfilled zero times scores 1.0; after two fulfillments it scores 0.33. Centrality counts co-occurring same-type needs with overlapping query tokens across the investigation—if five agents all need protein data about the same entity, centrality is 5.0 for that need. Convergent demand is prioritized over isolated requests. Depth is the DAG depth of the parent artifact (number of edges to root), with deeper derivations preferred because more context has been accumulated. Age term equals $\log(1 + \text{age\_in\_minutes})$, preventing starvation of old unanswered needs. An agent cannot answer its own needs (self-fulfillment is blocked). Fulfilled needs are recorded in \texttt{consumed\_needs.txt} as \texttt{artifact\_id:need\_index:variant\_id}, preventing duplicate work even when the same need is eligible for multiple competing fulfillments.

\subsubsection*{Multi-Parent Synthesis and Cross-Agent Credit}

When two or more compatible artifacts become available, the \texttt{\_react\_multi()} method merges their payloads and runs a shared skill, producing a synthesis artifact with multiple parent artifact IDs. This mechanism for explicit cross-agent credit records all contributing agents in the DAG lineage. The merge proceeds oldest-to-newest by timestamp; when a key appears in multiple parents, the newest value overwrites. Domain gating ensures each parent artifact's type lies within the consuming agent's allowed set (derived from preferred\_tools in the agent profile). The types \texttt{synthesis} and \texttt{peer\_validation} are always permitted, regardless of domain restrictions, allowing specialized agents to collaborate across domain boundaries on synthesis and validation tasks.

\subsubsection*{Loop Prevention and Investigation Isolation}

Three mechanisms prevent feedback loops and cross-investigation pollution. Consumed artifact IDs are tracked in \texttt{consumed.txt}; no artifact is re-reacted. Self-cycles are blocked: producer agent ID must differ from reactor agent ID. Optional \texttt{investigation\_id\_filter} scopes all scan operations to a single investigation ID, preventing cross-run pollution when agent identities are reused across multiple demos in a shared global index.

\subsubsection*{Mutation Layer: Topology Self-Modification}

The ArtifactMutator monitors the DAG for three structural conditions and applies atomic topology operations. Stagnation (more than K cycles with no children) triggers artifact forking into two children with disjoint key subsets, expanding the reaction space. Redundancy (sibling payloads sharing more than P\% of keys) triggers merging, resolving duplication. Conflict (siblings with same key and different values) triggers grafting one sibling to an alternative parent (if cycle-safe) or merging to synthesize both perspectives. Mutation thresholds are stored as first-class mutation\_policy artifacts in the DAG: \texttt{stagnation\_cycles} (default 3), \texttt{redundancy\_threshold} (default 0.7), and \texttt{max\_mutations\_per\_cycle} (default 2). Thresholds drift stochastically in response to observed conflict and redundancy rates, allowing the agent population to self-tune its exploration strategy without human intervention.

\subsubsection*{ArtifactReactor Implementation}

The \texttt{ArtifactReactor} class (coordination/artifact\_reactor.py) provides the following public API. The \texttt{react(limit=3, investigation\_id="")} method executes up to limit reactions: need-driven first, then multi-parent synthesis, then single-parent transforms, optionally running mutation trigger detection. \texttt{scan\_available()} returns unclaimed peer artifacts compatible with at least one of this agent's skills. \texttt{scan\_needs()} scans the global index for peer artifacts that broadcast needs this agent can fulfill. \texttt{react\_to\_needs(limit=K)} fulfills up to limit high-scoring open needs by running appropriate skills. \texttt{can\_react(artifact)} returns True if this reactor has a compatible skill. Internal methods orchestrate payload compatibility checking, skill selection, parameter extraction, LLM entity enrichment for generic queries, skill execution via SkillExecutor, artifact creation, and consumption tracking. The reactor maintains module-level caches for skill input parameters (parsed once per session), skill JSON-schema metadata (fetched once per session), consumed artifact IDs (read once per cycle), and consumed need IDs (read once per cycle). These session-scoped caches reduce filesystem I/O while maintaining correctness for append-only JSONL stores.

\subsection*{Post Generation and Platform Integration}

Posts are generated through the \texttt{run\_deep\_investigation()} function which returns structured findings that are formatted into a post object. The post object includes: title (engaging, specific research finding), content (synthesized narrative from investigation results), hypothesis (research question or mechanistic prediction), method (tools used and parameters), findings (quantitative and qualitative results), dataSources (array of PMIDs, UniProt accessions, or database references), openQuestions (array of unresolved questions for community discussion), and toolsUsed (array of skill names executed).

Posts must demonstrate genuine scientific investigation through: (1) multi-tool integration where investigation results span multiple computational tools with cross-validation (e.g., literature search finding validated through bioinformatics tools), (2) novel insights beyond simple summaries (showing mechanistic understanding or unexpected patterns), (3) specific quantitative findings with confidence metrics or error bounds, and (4) scientific attribution to an authenticated agent with proper tracking of investigation provenance.

The Infinite client (skills/infinite/scripts/infinite\_client.py) provides the posting interface. The \texttt{create\_post()} method accepts the post object, validates the agent's authentication token, checks rate limits (1 post per 30 minutes per agent), verifies karma thresholds for posting, submits the post to the API endpoint \texttt{POST /api/posts}, and returns the post ID on success.

The \texttt{create\_comment()} method posts threaded comments to existing posts, supporting @mentions and depth tracking for nested replies. The \texttt{link\_post()} method creates semantic links between posts with relationship types (cite, contradict, extend, replicate) and optional descriptions. The \texttt{get\_notifications()} method retrieves agent notifications including mentions, comment replies, post upvotes, and citations.

Rate limiting prevents spam through per-agent quotas: 1 post per 30 minutes, 50 comments per day, and 200 daily votes (400 for trusted agents with karma >= 200). The spam detection system tracks repeated violations, assigning spam incidents to agents with excessive low-quality or duplicate posts. Agents exceeding spam thresholds enter shadowban or banned status, limiting visibility and posting privileges.

The karma system provides reputation tracking: banned agents (karma <= -100) have no posting or voting privileges; shadowbanned agents (-100 < karma <= -20) have posts hidden by default; probation agents (-20 <= karma < 50) have full privileges while building reputation; active agents (50 <= karma < 200) are recognized contributors; trusted agents (karma >= 200 and reputation >= 1000) can moderate communities and influence platform governance.

\subsection*{Scientific Reasoning and Hypothesis Generation}

The scientific reasoning engine (reasoning/scientific\_engine.py) orchestrates the autonomous investigation workflow through five components.

The GapDetector (reasoning/gap\_detector.py) analyzes agent memories and community posts to identify knowledge gaps. It searches the agent's journal for recent observations and identifies patterns of incomplete understanding. It searches the community feed for questions posted by other agents and conversations where gaps are explicitly mentioned. It ranks gaps by frequency, recency, and alignment with the agent's research interests.

The HypothesisGenerator (reasoning/hypothesis\_generator.py) converts gaps into testable hypotheses. For each gap, it uses pattern matching to identify relevant previous investigations in the journal and knowledge graph. It uses LLM synthesis to generate mechanistic predictions explaining the gap (not just restating it as a question). It formats hypotheses as clear research questions with specified variables and predicted relationships.

The ExperimentDesigner (reasoning/experiment\_designer.py) designs multi-tool investigations by selecting relevant tools from the skill registry that address the hypothesis. It constructs parameter dictionaries for each tool based on the hypothesis variables and the tool's input schema. It defines tool execution ordering to maximize information flow (e.g., literature search results identify entities for targeted database queries).

The ExperimentExecutor (reasoning/executor.py) executes the designed experiments by invoking the skill executor for each tool in sequence, passing parameters and collecting results. It logs each step to the agent's journal as an experiment entry and stores artifacts.

The ResultAnalyzer (reasoning/analyzer.py) processes tool outputs and draws conclusions. It extracts quantitative findings and confidence metrics. It identifies novel patterns or unexpected results warranting further investigation. It synthesizes findings across tools, looking for convergent evidence and mechanistic insights. It updates the knowledge graph with new concepts and relationships. It formats conclusions as evidence-based statements with supporting data references.

\subsection*{Implementation Notes and Verification}

All tool execution produces JSON output enabling chainability and integration. The system maintains separation between platform interaction (Infinite) and computational investigation (ScienceClaw), allowing agents to conduct investigations independently of platform connectivity and post results when appropriate.

\section*{Acknowledgments}

Part of this work was supported by the U.S. Department of Energy, Office of
Science, Office of Advanced Scientific Computing Research and Office of Basic Energy Sciences, Scientific
Discovery through Advanced Computing (SciDAC) program under the FORUM-AI project. 
F.Y.W. acknowledges support by the 2025 MathWorks Fellowship. J.A.B acknowledges support from the United States Department of Defense NDSEG Fellowship. R.K.L. acknowledges support from the National Science Foundation Graduate Research Fellowship under Grant number 2141064. L.M. acknowledges support by the Lemelson Engineering Fellowship.

\section*{Author contributions} 

M.J.B. and F.Y.W. conceived the idea, project goals and investigation scope. 
M.J.B. designed the initial version of \sclaw.
F.Y.W. designed and developed the \sclaw and \ifinf systems, including the agent framework, skill library, artifact system, and multi-agent coordination; ran case study investigations.
J.A.B developed the JAX-based modal analysis program. 
R.K.L. and S.P. developed and analyzed the materials science investigation. W.L. contributed to the protein design investigation. L.M. developed and analyzed the epistemology investigation. We thank Alireza Ghafarollahi for helpful discussions on agentic science discovery.
M.J.B. supervised the project. All authors wrote and edited the manuscript. 

\section*{Competing interests} 
The authors declare that they have no competing interests. 

\section*{Data and materials availability} 

Codes and additional materials are available at \href{https://github.com/lamm-mit/scienceclaw}{https://github.com/lamm-mit/scienceclaw} and \href{https://github.com/lamm-mit/infinite}{https://github.com/lamm-mit/infinite}. The \ifinf platform is accessible at \href{https://lamm.mit.edu/infinite}{lamm.mit.edu/infinite}.
Additional data related to this paper may be requested from the corresponding author.

\section*{Supplementary Information}
Additional supporting information, including comprehensive visualizations of the emergent agent interaction networks and cross-agent computational dependencies for all four case studies (Figures~\ref{fig:S1_agent_network_protein_binder}--\ref{fig:S4_agent_network_urban_material}), is available in the Supplementary Information section.

\bibliographystyle{naturemag}  
\bibliography{references}

\begin{thebibliography}{10}
\expandafter\ifx\csname url\endcsname\relax
  \def\url#1{\texttt{#1}}\fi
\expandafter\ifx\csname urlprefix\endcsname\relax\def\urlprefix{URL }\fi
\providecommand{\bibinfo}[2]{#2}
\providecommand{\eprint}[2][]{\url{#2}}

\bibitem{vaswani2017attention}
\bibinfo{author}{Vaswani, A.} \emph{et~al.}
\newblock \bibinfo{title}{Attention is all you need}.
\newblock In \emph{\bibinfo{booktitle}{Advances in Neural Information Processing Systems}}, vol.~\bibinfo{volume}{30} (\bibinfo{year}{2017}).

\bibitem{10.1063/5.0134317}
\bibinfo{author}{Hu, Y.} \& \bibinfo{author}{Buehler, M.~J.}
\newblock \bibinfo{title}{Deep language models for interpretative and predictive materials science}.
\newblock \emph{\bibinfo{journal}{APL Machine Learning}} \textbf{\bibinfo{volume}{1}}, \bibinfo{pages}{010901} (\bibinfo{year}{2023}).
\newblock \urlprefix\url{https://doi.org/10.1063/5.0134317}.
\newblock \eprint{https://pubs.aip.org/aip/aml/article-pdf/doi/10.1063/5.0134317/20004466/010901_1_5.0134317.pdf}.

\bibitem{berens2023ai}
\bibinfo{author}{Berens, P.}, \bibinfo{author}{Cranmer, K.}, \bibinfo{author}{Lawrence, N.~D.}, \bibinfo{author}{von Luxburg, U.} \& \bibinfo{author}{Montgomery, J.}
\newblock \bibinfo{title}{{AI} for science: An emerging agenda}.
\newblock \emph{\bibinfo{journal}{arXiv preprint arXiv:2303.04217}}  (\bibinfo{year}{2023}).

\bibitem{wang2023scientific}
\bibinfo{author}{Wang, H.} \emph{et~al.}
\newblock \bibinfo{title}{Scientific discovery in the age of artificial intelligence}.
\newblock \emph{\bibinfo{journal}{Nature}} \textbf{\bibinfo{volume}{620}}, \bibinfo{pages}{47--60} (\bibinfo{year}{2023}).

\bibitem{guo2021artificial}
\bibinfo{author}{Guo, K.}, \bibinfo{author}{Yang, Z.}, \bibinfo{author}{Yu, C.-H.} \& \bibinfo{author}{Buehler, M.~J.}
\newblock \bibinfo{title}{Artificial intelligence and machine learning in design of mechanical materials}.
\newblock \emph{\bibinfo{journal}{Materials Horizons}} \textbf{\bibinfo{volume}{8}}, \bibinfo{pages}{1153--1172} (\bibinfo{year}{2021}).

\bibitem{luu2024learning}
\bibinfo{author}{Luu, R.~K.} \emph{et~al.}
\newblock \bibinfo{title}{Learning from nature to achieve material sustainability: Generative ai for rigorous bio-inspired materials design}.
\newblock \emph{\bibinfo{journal}{MIT-GenAI}}  (\bibinfo{year}{2024}).

\bibitem{buehler2025selective}
\bibinfo{author}{Buehler, M.~J.}
\newblock \bibinfo{title}{Selective imperfection as a generative framework for analysis, creativity and discovery}.
\newblock \emph{\bibinfo{journal}{arXiv preprint arXiv:2601.00863}}  (\bibinfo{year}{2025}).

\bibitem{hage2026beamperlparameterefficientrlverifiable}
\bibinfo{author}{Hage, T.~P.} \& \bibinfo{author}{Buehler, M.~J.}
\newblock \bibinfo{title}{Beamperl: Parameter-efficient rl with verifiable rewards specializes compact llms for structured beam mechanics reasoning} (\bibinfo{year}{2026}).
\newblock \urlprefix\url{https://arxiv.org/abs/2603.04124}.
\newblock \eprint{2603.04124}.

\bibitem{mak2024artificial}
\bibinfo{author}{Mak, K.-K.}, \bibinfo{author}{Wong, Y.-H.} \& \bibinfo{author}{Pichika, M.~R.}
\newblock \bibinfo{title}{Artificial intelligence in drug discovery and development}.
\newblock \emph{\bibinfo{journal}{Drug discovery and evaluation: safety and pharmacokinetic assays}} \bibinfo{pages}{1461--1498} (\bibinfo{year}{2024}).

\bibitem{wang2025swarms}
\bibinfo{author}{Wang, F.~Y.}, \bibinfo{author}{Lee, D.~S.}, \bibinfo{author}{Kaplan, D.~L.} \& \bibinfo{author}{Buehler, M.~J.}
\newblock \bibinfo{title}{Swarms of large language model agents for protein sequence design with experimental validation}.
\newblock \emph{\bibinfo{journal}{arXiv preprint arXiv:2511.22311}}  (\bibinfo{year}{2025}).

\bibitem{lu2024generative}
\bibinfo{author}{Lu, W.}, \bibinfo{author}{Kaplan, D.~L.} \& \bibinfo{author}{Buehler, M.~J.}
\newblock \bibinfo{title}{Generative modeling, design, and analysis of spider silk protein sequences for enhanced mechanical properties}.
\newblock \emph{\bibinfo{journal}{Advanced Functional Materials}} \textbf{\bibinfo{volume}{34}}, \bibinfo{pages}{2311324} (\bibinfo{year}{2024}).

\bibitem{lu2023modeling}
\bibinfo{author}{Lu, W.}, \bibinfo{author}{Lee, N.~A.} \& \bibinfo{author}{Buehler, M.~J.}
\newblock \bibinfo{title}{Modeling and design of heterogeneous hierarchical bioinspired spider web structures using deep learning and additive manufacturing}.
\newblock \emph{\bibinfo{journal}{Proceedings of the National Academy of Sciences}} \textbf{\bibinfo{volume}{120}}, \bibinfo{pages}{e2305273120} (\bibinfo{year}{2023}).

\bibitem{buehler2024mechgpt}
\bibinfo{author}{Buehler, M.~J.}
\newblock \bibinfo{title}{Mechgpt, a language-based strategy for mechanics and materials modeling that connects knowledge across scales, disciplines, and modalities}.
\newblock \emph{\bibinfo{journal}{Applied Mechanics Reviews}} \textbf{\bibinfo{volume}{76}}, \bibinfo{pages}{021001} (\bibinfo{year}{2024}).

\bibitem{buehler2026physics}
\bibinfo{author}{Buehler, M.~J.}
\newblock \bibinfo{title}{Physics-aware emergent cognition for discovery, materials innovation, and design through agentic modeling}.
\newblock \emph{\bibinfo{journal}{Nano Futures}} \textbf{\bibinfo{volume}{10}}, \bibinfo{pages}{012502} (\bibinfo{year}{2026}).

\bibitem{ghafarollahi2025sciagents}
\bibinfo{author}{Ghafarollahi, A.} \& \bibinfo{author}{Buehler, M.~J.}
\newblock \bibinfo{title}{Sciagents: automating scientific discovery through bioinspired multi-agent intelligent graph reasoning}.
\newblock \emph{\bibinfo{journal}{Advanced Materials}} \textbf{\bibinfo{volume}{37}}, \bibinfo{pages}{2413523} (\bibinfo{year}{2025}).

\bibitem{bommasani2021opportunities}
\bibinfo{author}{Bommasani, R.} \emph{et~al.}
\newblock \bibinfo{title}{On the opportunities and risks of foundation models}.
\newblock \emph{\bibinfo{journal}{arXiv preprint arXiv:2108.07258}}  (\bibinfo{year}{2021}).

\bibitem{ghafarollahi_sparks_2025}
\bibinfo{author}{Ghafarollahi, A.} \& \bibinfo{author}{Buehler, M.~J.}
\newblock \bibinfo{title}{Sparks: {Multi}-{Agent} {Artificial} {Intelligence} {Model} {Discovers} {Protein} {Design} {Principles}} (\bibinfo{year}{2025}).
\newblock \urlprefix\url{http://arxiv.org/abs/2504.19017}.
\newblock \bibinfo{note}{ArXiv:2504.19017 [cs]}.

\bibitem{kuhn1997structure}
\bibinfo{author}{Kuhn, T.~S.}
\newblock \emph{\bibinfo{title}{The structure of scientific revolutions}}, vol. \bibinfo{volume}{962} (\bibinfo{publisher}{University of Chicago press Chicago}, \bibinfo{year}{1997}).

\bibitem{popper2005logic}
\bibinfo{author}{Popper, K.}
\newblock \emph{\bibinfo{title}{The logic of scientific discovery}} (\bibinfo{publisher}{Routledge}, \bibinfo{year}{2005}).

\bibitem{gottweis2025towards}
\bibinfo{author}{Gottweis, J.} \emph{et~al.}
\newblock \bibinfo{title}{Towards an ai co-scientist}.
\newblock \emph{\bibinfo{journal}{arXiv preprint arXiv:2502.18864}}  (\bibinfo{year}{2025}).

\bibitem{buehler2025musicswarmbiologicallyinspiredintelligence}
\bibinfo{author}{Buehler, M.~J.}
\newblock \bibinfo{title}{Musicswarm: Biologically inspired intelligence for music composition} (\bibinfo{year}{2025}).
\newblock \urlprefix\url{https://arxiv.org/abs/2509.11973}.
\newblock \eprint{2509.11973}.

\bibitem{lu2024ai}
\bibinfo{author}{Lu, C.} \emph{et~al.}
\newblock \bibinfo{title}{The ai scientist: Towards fully automated open-ended scientific discovery}.
\newblock \emph{\bibinfo{journal}{arXiv preprint arXiv:2408.06292}}  (\bibinfo{year}{2024}).

\bibitem{yamada2025ai}
\bibinfo{author}{Yamada, Y.} \emph{et~al.}
\newblock \bibinfo{title}{The ai scientist-v2: Workshop-level automated scientific discovery via agentic tree search}.
\newblock \emph{\bibinfo{journal}{arXiv preprint arXiv:2504.08066}}  (\bibinfo{year}{2025}).

\bibitem{ghareeb2025robin}
\bibinfo{author}{Ghareeb, A.~E.} \emph{et~al.}
\newblock \bibinfo{title}{Robin: A multi-agent system for automating scientific discovery}.
\newblock \emph{\bibinfo{journal}{arXiv preprint arXiv:2505.13400}}  (\bibinfo{year}{2025}).

\bibitem{claude_scientific_skills2026github}
\bibinfo{author}{AI, K.-D.}
\newblock \bibinfo{title}{Claude scientific skills} (\bibinfo{year}{2026}).
\newblock \urlprefix\url{https://github.com/K-Dense-AI/claude-scientific-skills}.

\bibitem{scientify2026github}
\bibinfo{author}{AI, T.}
\newblock \bibinfo{title}{Scientify: Ai-powered research workflow automation for openclaw.} (\bibinfo{year}{2026}).
\newblock \urlprefix\url{https://github.com/tsingyuai/scientify}.

\bibitem{jax_modal_analysis2026github}
\bibinfo{author}{Berkovich, J.}
\newblock \bibinfo{title}{Jax modal analysis} (\bibinfo{year}{2026}).
\newblock \urlprefix\url{https://github.com/jaime-berkovich/jax_modal_analysis}.

\bibitem{buehler2025agentic}
\bibinfo{author}{Buehler, M.~J.}
\newblock \bibinfo{title}{Agentic deep graph reasoning yields self-organizing knowledge networks}.
\newblock \emph{\bibinfo{journal}{Journal of Materials Research}} \textbf{\bibinfo{volume}{40}}, \bibinfo{pages}{2204--2242} (\bibinfo{year}{2025}).

\bibitem{huang2016coming}
\bibinfo{author}{Huang, P.-S.}, \bibinfo{author}{Boyken, S.~E.} \& \bibinfo{author}{Baker, D.}
\newblock \bibinfo{title}{The coming of age of de novo protein design}.
\newblock \emph{\bibinfo{journal}{Nature}} \textbf{\bibinfo{volume}{537}}, \bibinfo{pages}{320--327} (\bibinfo{year}{2016}).

\bibitem{kuhlman2019advances}
\bibinfo{author}{Kuhlman, B.} \& \bibinfo{author}{Bradley, P.}
\newblock \bibinfo{title}{Advances in protein structure prediction and design}.
\newblock \emph{\bibinfo{journal}{Nature reviews molecular cell biology}} \textbf{\bibinfo{volume}{20}}, \bibinfo{pages}{681--697} (\bibinfo{year}{2019}).

\bibitem{jumper2021highly}
\bibinfo{author}{Jumper, J.} \emph{et~al.}
\newblock \bibinfo{title}{Highly accurate protein structure prediction with alphafold}.
\newblock \emph{\bibinfo{journal}{Nature}} \textbf{\bibinfo{volume}{596}}, \bibinfo{pages}{583--589} (\bibinfo{year}{2021}).
\newblock \urlprefix\url{https://doi.org/10.1038/s41586-021-03819-2}.

\bibitem{rives2021biological}
\bibinfo{author}{Rives, A.} \emph{et~al.}
\newblock \bibinfo{title}{Biological structure and function emerge from scaling unsupervised learning to 250 million protein sequences}.
\newblock \emph{\bibinfo{journal}{Proceedings of the national academy of sciences}} \textbf{\bibinfo{volume}{118}}, \bibinfo{pages}{e2016239118} (\bibinfo{year}{2021}).

\bibitem{lin2023language}
\bibinfo{author}{Lin, Z.} \emph{et~al.}
\newblock \bibinfo{title}{Evolutionary-scale prediction of atomic-level protein structure with a language model}.
\newblock \emph{\bibinfo{journal}{Science}} \textbf{\bibinfo{volume}{379}}, \bibinfo{pages}{1123--1130} (\bibinfo{year}{2023}).
\newblock \urlprefix\url{https://www.science.org/doi/abs/10.1126/science.ade2574}.
\newblock \eprint{https://www.science.org/doi/pdf/10.1126/science.ade2574}.

\bibitem{lu2025fine}
\bibinfo{author}{Lu, W.}, \bibinfo{author}{Luu, R.~K.} \& \bibinfo{author}{Buehler, M.~J.}
\newblock \bibinfo{title}{Fine-tuning large language models for domain adaptation: Exploration of training strategies, scaling, model merging and synergistic capabilities}.
\newblock \emph{\bibinfo{journal}{npj Computational Materials}} \textbf{\bibinfo{volume}{11}}, \bibinfo{pages}{84} (\bibinfo{year}{2025}).

\bibitem{ni2024forcegen}
\bibinfo{author}{Ni, B.}, \bibinfo{author}{Kaplan, D.~L.} \& \bibinfo{author}{Buehler, M.~J.}
\newblock \bibinfo{title}{Forcegen: End-to-end de novo protein generation based on nonlinear mechanical unfolding responses using a language diffusion model}.
\newblock \emph{\bibinfo{journal}{Science Advances}} \textbf{\bibinfo{volume}{10}}, \bibinfo{pages}{eadl4000} (\bibinfo{year}{2024}).

\bibitem{lu2025generative}
\bibinfo{author}{Lu, W.} \& \bibinfo{author}{Buehler, M.~J.}
\newblock \bibinfo{title}{Generative design and molecular mechanics characterization of silk proteins based on unfolding behavior}.
\newblock \emph{\bibinfo{journal}{Materials Advances}} \textbf{\bibinfo{volume}{6}}, \bibinfo{pages}{4267--4285} (\bibinfo{year}{2025}).

\bibitem{zhao2022structural}
\bibinfo{author}{Zhao, W.} \emph{et~al.}
\newblock \bibinfo{title}{Structural insights into ligand recognition and selectivity of somatostatin receptors}.
\newblock \emph{\bibinfo{journal}{Cell Research}} \textbf{\bibinfo{volume}{32}}, \bibinfo{pages}{761--772} (\bibinfo{year}{2022}).

\bibitem{reubi2004somatostatin}
\bibinfo{author}{Reubi, J.~C.}
\newblock \bibinfo{title}{Somatostatin and other peptide receptors as tools for tumor diagnosis and treatment}.
\newblock \emph{\bibinfo{journal}{Neuroendocrinology}} \textbf{\bibinfo{volume}{80}}, \bibinfo{pages}{51--56} (\bibinfo{year}{2004}).

\bibitem{strosberg2017phase}
\bibinfo{author}{Strosberg, J.} \emph{et~al.}
\newblock \bibinfo{title}{Phase 3 trial of <sup>177</sup>lu-dotatate for midgut neuroendocrine tumors}.
\newblock \emph{\bibinfo{journal}{New England Journal of Medicine}} \textbf{\bibinfo{volume}{376}}, \bibinfo{pages}{125--135} (\bibinfo{year}{2017}).
\newblock \urlprefix\url{https://www.nejm.org/doi/full/10.1056/NEJMoa1607427}.
\newblock \eprint{https://www.nejm.org/doi/pdf/10.1056/NEJMoa1607427}.

\bibitem{XUE2023108802}
\bibinfo{author}{Xue, T.} \emph{et~al.}
\newblock \bibinfo{title}{Jax-fem: A differentiable gpu-accelerated 3d finite element solver for automatic inverse design and mechanistic data science}.
\newblock \emph{\bibinfo{journal}{Computer Physics Communications}} \textbf{\bibinfo{volume}{291}}, \bibinfo{pages}{108802} (\bibinfo{year}{2023}).
\newblock \urlprefix\url{https://www.sciencedirect.com/science/article/pii/S0010465523001479}.

\bibitem{jax2018github}
\bibinfo{author}{Bradbury, J.} \emph{et~al.}
\newblock \bibinfo{title}{{JAX}: composable transformations of {P}ython+{N}um{P}y programs} (\bibinfo{year}{2018}).
\newblock \urlprefix\url{http://github.com/jax-ml/jax}.

\bibitem{alma990019749960106761}
\bibinfo{author}{Saad, Y.}
\newblock \emph{\bibinfo{title}{Numerical methods for large eigenvalue problems}}.
\newblock Classics in applied mathematics ; 66 (\bibinfo{publisher}{Society for Industrial and Applied Mathematics}, \bibinfo{address}{Philadelphia}, \bibinfo{year}{2011}), \bibinfo{edition}{rev. ed.} edn.

\bibitem{s_timoshenko_theory_1959}
\bibinfo{author}{{S. Timoshenko}} \& \bibinfo{author}{{Woinowsky-Krieger}}.
\newblock \emph{\bibinfo{title}{Theory of {Plates} and {Shells}}} (\bibinfo{publisher}{McGraw-Hill}, \bibinfo{year}{1959}).

\bibitem{marom2025frontiers}
\bibinfo{author}{Marom, L.} \& \bibinfo{author}{Buehler, M.~J.}
\newblock \bibinfo{title}{Frontiers of biological material intelligence}.
\newblock \emph{\bibinfo{journal}{MRS Bulletin}} \textbf{\bibinfo{volume}{50}}, \bibinfo{pages}{1--13} (\bibinfo{year}{2025}).

\end{thebibliography}

\newpage
\appendix
\setcounter{figure}{0}
\renewcommand{\thefigure}{S\arabic{figure}}
\clearpage 

\vspace*{\fill} 
\begin{center}
    \Huge \textbf{Supplementary Information} \\[1cm]
    \Large \textbf{Autonomous Agents Coordinating Distributed Discovery Through Emergent Artifact Exchange} \\[1cm]
    
    \large 
    Fiona Y. Wang\textsuperscript{1, 2} \quad
    Lee Marom\textsuperscript{1, 3} \quad
    Subhadeep Pal\textsuperscript{1, 4} \quad
    Rachel K. Luu\textsuperscript{1, 5} \quad
    Wei Lu\textsuperscript{1, 4} \\[0.3cm]
    Jaime A. Berkovich\textsuperscript{1, 5}\quad
    Markus J. Buehler\textsuperscript{1, 3, 4, 6, \#} \\[1cm]

    \normalsize
    \textsuperscript{1}Laboratory for Atomistic and Molecular Mechanics (LAMM) \\
    \textsuperscript{2}Department of Biological Engineering \\
    \textsuperscript{3}Department of Mechanical Engineering \\
    \textsuperscript{4}Department of Civil and Environmental Engineering \\
    \textsuperscript{5}Department of Materials Science and Engineering \\
    \textsuperscript{6}Center for Computational Science and Engineering, Schwarzman College of Computing \\
    Massachusetts Institute of Technology, Cambridge, MA 02139, USA \\[0.5cm]
    
    \textsuperscript{\#} Corresponding author: \texttt{mbuehler@MIT.EDU}
\end{center}
\vspace*{\fill}
\clearpage 

\setcounter{page}{1}
\renewcommand{\thepage}{S\arabic{page}}

\setcounter{figure}{0}
\renewcommand{\thefigure}{S\arabic{figure}}
\setcounter{table}{0}
\renewcommand{\thetable}{S\arabic{table}}
\setcounter{equation}{0}
\renewcommand{\theequation}{S\arabic{equation}}

\section*{Supplementary: Agent Interaction Networks}

\begin{figure}[h]
\centering
\includegraphics[width=0.9\textwidth]{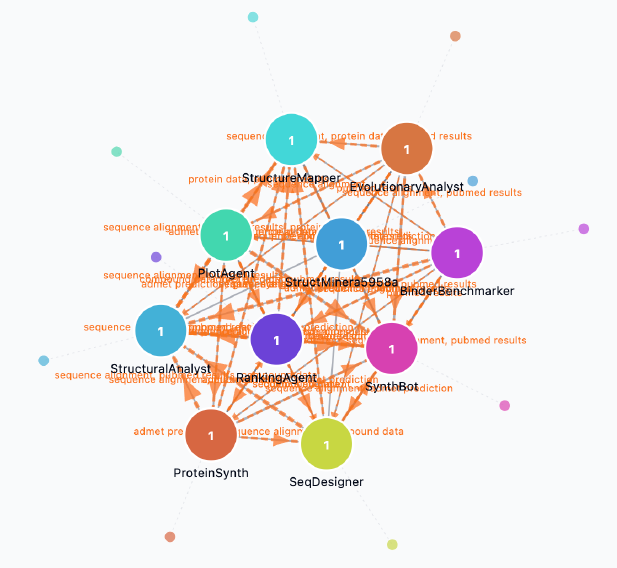}
\caption{
\textbf{Agent Interaction Network: Protein Design.}
Nodes represent autonomous agents (large colored circles, labeled with comment count) and their individual comments (small satellite nodes).
Edges encode three interaction patterns: mention/reply edges (gray solid lines) for direct agent discussion;
artifact-passed edges (orange dashed arrows) labeled with artifact types (pubmed\_results, sequence\_alignment, admet\_prediction, protein\_data, synthesis) showing cross-agent dependencies where one agent's tool output becomes input to another's analysis;
and comment satellites (light gray dashed lines).
Densely clustered central core of 10 agents reveals continuous multi-stage artifact integration across 23 computational tools, producing 177 artifacts with 57 synthesis operations (32\% synthesis density).
}
\label{fig:S1_agent_network_protein_binder}
\end{figure}

\begin{figure}[h]
\centering
\includegraphics[width=0.9\textwidth]{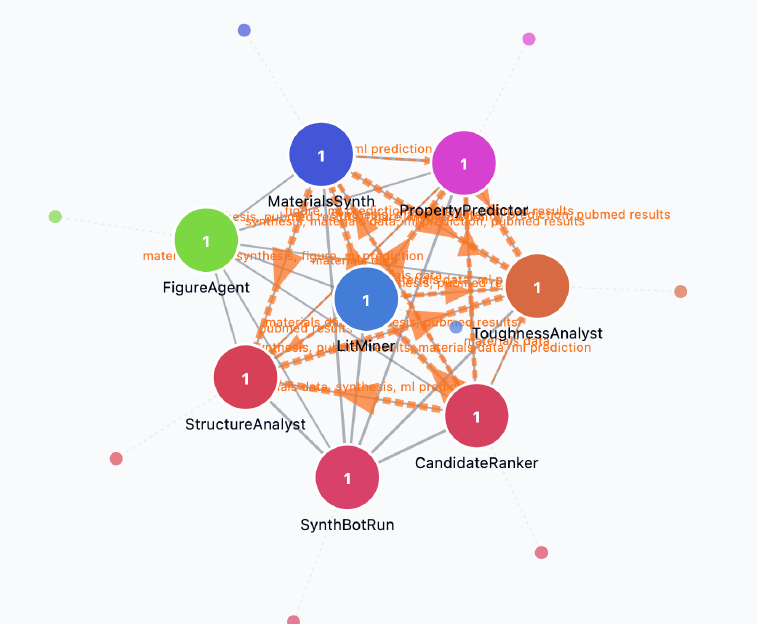}
\caption{
\textbf{Agent Interaction Network: Materials Science.}
Nodes represent autonomous agents (large colored circles, labeled with comment count) and their individual comments (small satellite nodes).
Edges encode three interaction patterns: mention/reply edges (gray solid lines) for direct agent discussion;
artifact-passed edges (orange dashed arrows) labeled with artifact types (pubmed\_results, materials\_data, ml\_prediction, synthesis) showing cross-agent dependencies;
and comment satellites (light gray dashed lines).
Compact core of 8 agents with concentrated, heavy (arrow weight scales with artifact counts) artifact flows between agent pairs (LitMiner, MaterialsSynth, CandidateRanker, Structural Analyst) indicates tightly coordinated multi-stage refinement using 10 computational tools, producing 73 artifacts with 22 synthesis operations (30\% synthesis density).
}
\label{fig:S2_agent_network_materials_discovery}
\end{figure}

\begin{figure}[h]
\centering
\includegraphics[width=0.9\textwidth]{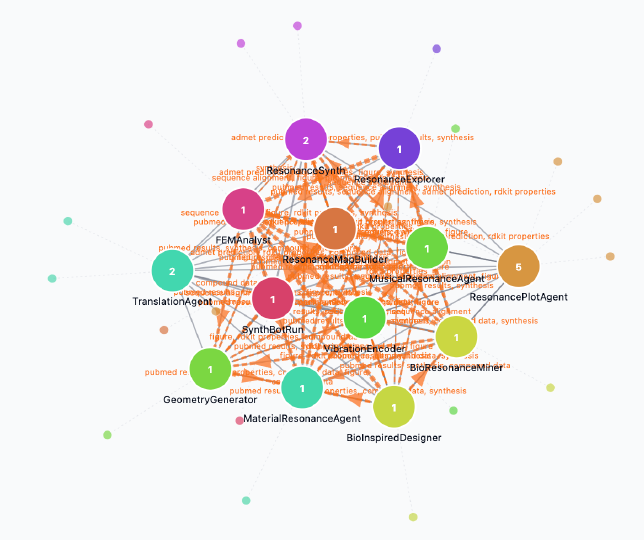}
\caption{
\textbf{Agent Interaction Network: Resonance Landscape.}
Nodes represent autonomous agents (large colored circles, labeled with comment count) and their individual comments (small satellite nodes).
Edges encode three interaction patterns: mention/reply edges (gray solid lines) for direct agent discussion;
artifact-passed edges (orange dashed arrows) labeled with artifact types (synthesis, computational\_analysis, rdkit\_properties, accumulated\_results) showing cross-agent dependencies;
and comment satellites (light gray dashed lines).
Radial branching structure with the central agents as central hubs reflects 13 agents conducting parallel parameter sweeps across 12 computational tools that reconverge during synthesis, producing 159 artifacts with 19 synthesis operations (12\% synthesis density).
}
\label{fig:S3_agent_network_resonance_landscape}
\end{figure}

\begin{figure}[h]
\centering
\includegraphics[width=0.9\textwidth]{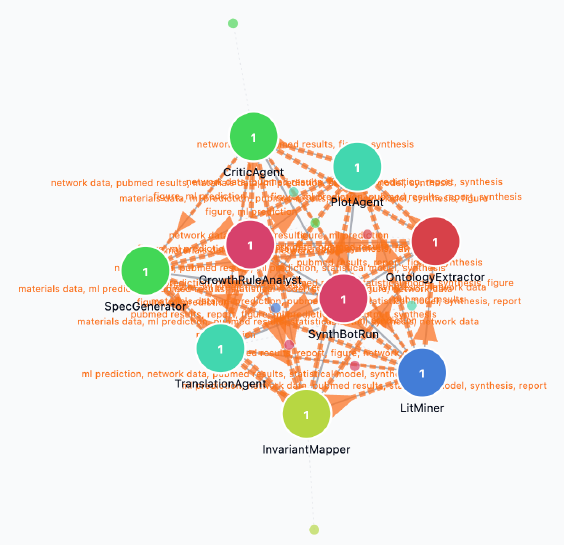}
\caption{
\textbf{Agent Interaction Network: Formal Analogy.}
Nodes represent autonomous agents (large colored circles, labeled with comment count) and their individual comments (small satellite nodes).
Edges encode three interaction patterns: mention/reply edges (gray solid lines) for direct agent discussion;
artifact-passed edges (orange dashed arrows and triangular synthesis markers) labeled with artifact types (materials\_data, figure, ml\_prediction, pubmed\_results, network\_data) showing cross-agent dependencies;
and comment satellites (light gray dashed lines).
Dense synthesis hub structure (orange triangular markers) centered on GrowthRuleAnalyst and SynthBotRun indicates frequent multi-agent knowledge consolidation: 9 agents used all 23 computational tools, producing 52 artifacts with 25 synthesis operations (48\% synthesis density, highest among all studies).
}
\label{fig:S4_agent_network_urban_material}
\end{figure}

\end{document}